\documentclass[11pt]{article}

\usepackage[final]{acl}

\usepackage{times}
\usepackage{latexsym}
\usepackage[T1]{fontenc}
\usepackage[utf8]{inputenc}
\usepackage{microtype}
\usepackage{inconsolata}
\usepackage{graphicx}
\usepackage{url}
\usepackage{colortbl}
\usepackage{multirow}
\usepackage{amsmath}

\title{mdok-style at SemEval-2026 Task 9: Finetuning LLMs for Multilingual Polarization Detection}

\author{Dominik Macko$^1$, Alok Debnath$^2$, Jakub Simko$^1$\\
  $^{1}$ Kempelen Institute of Intelligent Technologies, Bratislava, Slovakia\\
  $^{2}$ ADAPT Centre, School of Computer Science and Statistics, Trinity College Dublin\\
 \texttt{\{dominik.macko, jakub.simko\}}@kinit.sk, \texttt{\{alok.debnath\}}@adaptcentre.ie \\}

\begin{document}
\maketitle
\begin{abstract}
SemEval-2026 Task~9 is focused on multilingual polarization detection. Specifically, it covers the identification of multilingual, multicultural and multievent polarization along three axes (in subtasks), namely detection, type, and manifestation. Online polarization presents a concern, because it is often followed by hate speech, offensive discourse, and social fragmentation. Therefore, its detection before it escalates is crucial for a safer and more inclusive online space. We have coped with this SemEval task by finetuning mid-size LLMs for the sequence-classification task using the QLoRA parameter-efficient finetuning technique. The training data augmented the multilingual (22 languages) training sets by anonymized, lower-cased, upper-cased, and homoglyphied counterparts, making the detection more robust.
\end{abstract}

\section{Introduction}

Automated detection of online polarization is an important problem, as it could help to mitigate polarization before it escalates. This challenge is dealt with by the POLAR shared task~\citep{naseem2026polarbenchmarkmultilingualmulticultural}, officially SemEval-2026 Task~9. Within its three subtasks, it is aimed at binary polarization detection (subtask 1, determining whether a given text expresses polarization), polarization type classification (subtask 2, identifying the social dimension underlying the polarization), and manifestation identification (subtask 3, detecting how polarization is rhetorically manifested). All of them are handled multilingually, covering 22 languages.

The submitted multilingual system\footnote{\scriptsize\url{https://github.com/kinit-sk/mdok-style-polar2026}} is heavily based on mdok (\textbf{m}achine \textbf{d}etector \textbf{o}f \textbf{K}InIT), a robust detector of machine-generated text presented at PAN@CLEF2025~\citep{Bevendorff2025OverviewOT}. Besides a simple transferring of the existing system to a new task, we also explore possibilities of using appraisal-theoretic\footnote{appraisals are cognitive signals of emotion} detection of emotions (including various appraisal dimensions, e.g., pleasantness, self-control, or suddenness)~\citep{hofmann-etal-2020-appraisal,debnath-etal-2025-appraisal}, for the tasks of polarization detection.

\section{Background}

We have utilized our experience in a binary machine-generated text detection task and transferred our best approach to polarization detection task.
Our first experience with finetuning LLMs for a binary classification task was at SemEval-2024 Task~8~\citep{spiegel-macko-2024-kinit}, where we have identified better detection performance achievable by finetuned a 7B-parameters LLM than the traditional small pre-trained models (BERT-like). We have further explored and increased the robustness of the finetuning process~\citep{macko2025increasingrobustnessfinetunedmultilingual}, and eventually resulted into the mdok~\citep{mdok} finetuning approach, ranking 1st in both subtasks of PAN@CLEF2025~\citep{Bevendorff2025OverviewOT}. In one of the subtasks, mdok was extended to the multiclass scenario (hybrid human-AI collaboration) and further modified to multilingual authorship attribution~\citep{lacava2025authorshipattributionmultilingualmachinegenerated}. In this SemEval-2026 Task~9 shared task~\citep{naseem-etal-2026-polar}, we have modified the multiclass mdok version to the multiclass multi-label classification required in the subtasks 2 and 3.

\section{System Overview}

The proposed system contains multiple components that are described below. The appraisal annotation has not been used in the submitted system (due to timing); however, we include it as an alternative we have explored.

\subsection{Data Augmentation}

To increase the number of training samples and to make the detection more robust, we augment the original training data. In our approach for binary polarization detection (subtask 1), we use various augmentation techniques described below. Essentially, for each technique, the training texts are duplicated and the duplicated part is modified by the corresponding technique. Afterwards de-duplication takes care about the redundant (unmodified) texts.

\paragraph{Anonymization.}
This text-prepocessing procedure replaces identified email addresses, user mentions, and phone numbers by the tags of [EMAIL], [USER], [PHONE]. Such an anonymization approach reduces biases towards specific personal information, which could appear in the training data and focuses more on the meaning of the text. We use this technique as a data augmentation, duplicating the original texts, preprocessing by this anonymization procedure and then combining with the original texts.

\paragraph{Lower-casing and Upper-casing.}
To further make the detection more sensitive on actual meaning than on the visual appearance of the texts (especially important on social media using wilder style of communication), we convert the duplicated texts to all lowercase, and also to all uppercase. This effectively increases the training size, making the decision of the detection case insensitive.

\paragraph{Homoglyphication.}
In our previous work on machine-generated text detection, we have identified homoglyph attack (replacing some characters to visually similar counterparts of another script) as making one of the highest confusion on text classifiers (disrupting their tokenization). Such modification causes detection evasion in many cases; however, can be effectively dealt with when considered during training. We have applied such homoglyph attack to the duplicated portion of the training texts, making the polarization detection more robust against obfuscation.

\subsection{Finetuning Process}

For finetuning the multilingual LLM for polarization task, we have combined the train samples of all languages, which effectively increases the train samples count and enhances the focus on the task rather than on the language.

Since the provided development-set samples are imbalanced across languages, we have combined the provided development data with the training data and de-duplicated such a train set. Afterwards, for subtask 1, we have sampled 100 texts per each language and label (i.e., 200 samples per language $\times$ 22 languages) resulting in 4400 samples for validation during the finetuning. For subtasks 2 and 3, we have just pseudorandomly sampled 100 texts per language, reflecting multi-labels distribution of the train set (i.e., not perfectly balanced).

In subtask 1, we have increased the train set by above mentioned four augmentation technique by 20\% (i.e., 5\% each). It makes the trained classifier more robust (invariant to the visual form of the text), focusing on the meaning rather than on representation.

For all subtasks, we have used QLoRA~\citep{NEURIPS2023_1feb8787} parameter-efficient finetuning (PEFT) approach with 4-bit quantization using bitsandbytes library within HuggingFace transformers\footnote{\url{https://github.com/huggingface/transformers}} python framework. Regarding hyperparameters, we have applied constant learning rate of 2e-5, a warmup ratio of 0.03, paged Adamoptimizer, a batch size of 1 sample, and validation each 500 steps. The finetuning process has taken a single epoch with the final checkpoint selection based on the best metric achieved on validation set (AUC\footnote{area under curve of the receiver operating characteristic} for subtask 1, macro avg. F1 for subtasks 2 and 3).

We have published the full source code to retrain the detectors (see the footnote on the first page), making the detectors replicable. To install the dependencies, just clone the IMGTB framework~\citep{spiegel-macko-2024-imgtb} repository\footnote{\url{https://github.com/kinit-sk/IMGTB}}, install the corresponding conda environment, and update the transformers library (for support of the newest models).

\subsection{Base Model Selection}

We have focused on a single model for all the languages (22 in total) combined; therefore, we have limited the base-model selection to multilingual models that can cope with low-resource languages as well.
For subtask 1, we have experimented with multiple models, while Qwen3-32B~\citep{yang2025qwen3technicalreport} (support of >100 languages) has been selected to be the best performant. For subtasks 2 and 3, we had problems with multi-label classification finetuning of Qwen3 models; therefore, we have resulted with Gemma-3-27B-pt~\citep{gemmateam2025gemma3technicalreport} (support of >140 languages) in these subtasks.

\subsection{Appraisal Annotation}

Appraisal theory is an evolutionary theory of emotion that suggests that emotions are outcomes of how individuals ``appraise'' or value an event or experience on cognitive dimensions such as its \textit{pleasantness}, \textit{predictability}, and the \textit{control} the experiencer has on the situation.

The ISEAR corpus is a benchmark dataset of categorical emotion annotations for private perspectives on emotional scenarios to understand the causes and reactions of emotions as represented in text \citep{scherer1994evidence}. 
For multilingual appraisal analysis, the \textbf{enISEAR} and \textbf{deISEAR} datasets were used \citep{troiano-etal-2019-crowdsourcing}. 
Both corpora provide crowd-sourced classification annotations for seven emotion categories (anger, disgust, fear, sadness, shame, joy, and guilt), as well as binary ratings for five appraisal categories (consequences to self, consequences to others, degree of control, degree of responsibility, alignment with social values), and four event descriptions (general, past, future, or prospective).
Both deISEAR and enISEAR comprise 1001 event-centered descriptions: deISEAR includes 1084 sentences, while enISEAR contains 1366 sentences.

Appraisal estimation was treated as a multitask regression task, with an architecture similar to AppraisePLM \citep{debnath-etal-2025-appraisal}, except that for handling multilingual inputs, the model uses an XLM-Roberta \citep{conneau2020unsupervised} encoder representation for text embeddings. The loss function is also adjusted to account for the binary nature of the appraisal and event description categories, using a weighted combination of MSE and BCE loss. 

A more detailed description of appraisal theory, the appraisal dimensions used for analysis, and the AppraisePLM architecture are provided in Appendix~\ref{sec:B}.


\section{Experimental Setup}

In each subtask, the submitted systems used the official train and development sets for training as mentioned above and the official test set for testing.

The official metric for all subtasks is the macro average of per-class F1 scores (F1 is a harmonic mean of precision and recall), separately for each language.
Percentile value is calculated as a number of systems ranked worse than our system our of all systems.

For exploration of usage of appraisals annotations for polarization detection, we have annotated only the train set (due to time constraints), which has been further split for training (80\% of data) and testing (20\% of data) of the LogisticRegression classifier with default parameters (random state of 42). For each language, a separate classifier has been trained. The results of predictions has been evaluated per subtask also using macro F1 metric.

\section{Results}

Performance of the submitted system (trained as described above) in a form of macro avg. F1 (Macro F1) is provided in Table~\ref{tab:performance}. Darker color of the background gradient reflects higher performance. Four languages without the performance value (-) have not been included in subtask 3.

\begin{table}[!b]
\centering
\resizebox{0.95\linewidth}{!}{
\begin{tabular}{r|c|c|c}
\hline
 & \multicolumn{3}{c}{\textbf{Macro F1}}\\
\bfseries Language & \bfseries Subtask 1 & \bfseries Subtask 2 & \bfseries Subtask 3 \\
\hline
\bfseries amh & {\cellcolor[HTML]{ADC1DD}} \color[HTML]{000000} 0.6619 & {\cellcolor[HTML]{D2D3E7}} \color[HTML]{000000} 0.5116 & {\cellcolor[HTML]{E0DEED}} \color[HTML]{000000} 0.4310 \\
\bfseries arb & {\cellcolor[HTML]{78ABD0}} \color[HTML]{000000} 0.8348 & {\cellcolor[HTML]{B5C4DF}} \color[HTML]{000000} 0.6279 & {\cellcolor[HTML]{D2D2E7}} \color[HTML]{000000} 0.5157 \\
\bfseries ben & {\cellcolor[HTML]{76AAD0}} \color[HTML]{000000} 0.8415 & {\cellcolor[HTML]{F3EDF5}} \color[HTML]{000000} 0.3050 & {\cellcolor[HTML]{FFF7FB}} \color[HTML]{000000} 0.1272 \\
\bfseries deu & {\cellcolor[HTML]{96B6D7}} \color[HTML]{000000} 0.7398 & {\cellcolor[HTML]{CDD0E5}} \color[HTML]{000000} 0.5399 & {\cellcolor[HTML]{E5E1EF}} \color[HTML]{000000} 0.4044 \\
\bfseries eng & {\cellcolor[HTML]{81AED2}} \color[HTML]{000000} 0.8058 & {\cellcolor[HTML]{DDDBEC}} \color[HTML]{000000} 0.4519 & {\cellcolor[HTML]{EBE6F2}} \color[HTML]{000000} 0.3697 \\
\bfseries fas & {\cellcolor[HTML]{8CB3D5}} \color[HTML]{000000} 0.7690 & {\cellcolor[HTML]{D0D1E6}} \color[HTML]{000000} 0.5250 & {\cellcolor[HTML]{F1EBF5}} \color[HTML]{000000} 0.3208 \\
\bfseries hau & {\cellcolor[HTML]{96B6D7}} \color[HTML]{000000} 0.7401 & {\cellcolor[HTML]{FFF7FB}} \color[HTML]{000000} 0.1689 & {\cellcolor[HTML]{FFF7FB}} \color[HTML]{000000} 0.0000 \\
\bfseries hin & {\cellcolor[HTML]{84B0D3}} \color[HTML]{000000} 0.7974 & {\cellcolor[HTML]{91B5D6}} \color[HTML]{000000} 0.7573 & {\cellcolor[HTML]{94B6D7}} \color[HTML]{000000} 0.7453 \\
\bfseries ita & {\cellcolor[HTML]{99B8D8}} \color[HTML]{000000} 0.7303 & {\cellcolor[HTML]{F3EDF5}} \color[HTML]{000000} 0.3019 & {\cellcolor[HTML]{FFF7FB}} \color[HTML]{000000} - \\
\bfseries khm & {\cellcolor[HTML]{B5C4DF}} \color[HTML]{000000} 0.6293 & {\cellcolor[HTML]{B4C4DF}} \color[HTML]{000000} 0.6323 & {\cellcolor[HTML]{FAF2F8}} \color[HTML]{000000} 0.2482 \\
\bfseries mya & {\cellcolor[HTML]{69A5CC}} \color[HTML]{000000} 0.8788 & {\cellcolor[HTML]{A7BDDB}} \color[HTML]{000000} 0.6835 & {\cellcolor[HTML]{FFF7FB}} \color[HTML]{000000} - \\
\bfseries nep & {\cellcolor[HTML]{63A2CB}} \color[HTML]{000000} 0.8915 & {\cellcolor[HTML]{83AFD3}} \color[HTML]{000000} 0.8026 & {\cellcolor[HTML]{C5CCE3}} \color[HTML]{000000} 0.5669 \\
\bfseries ori & {\cellcolor[HTML]{83AFD3}} \color[HTML]{000000} 0.8013 & {\cellcolor[HTML]{DBDAEB}} \color[HTML]{000000} 0.4628 & {\cellcolor[HTML]{FFF7FB}} \color[HTML]{000000} 0.1227 \\
\bfseries pan & {\cellcolor[HTML]{8CB3D5}} \color[HTML]{000000} 0.7736 & {\cellcolor[HTML]{EAE6F1}} \color[HTML]{000000} 0.3734 & {\cellcolor[HTML]{D6D6E9}} \color[HTML]{000000} 0.4933 \\
\bfseries pol & {\cellcolor[HTML]{7EADD1}} \color[HTML]{000000} 0.8158 & {\cellcolor[HTML]{CED0E6}} \color[HTML]{000000} 0.5350 & {\cellcolor[HTML]{FFF7FB}} \color[HTML]{000000} - \\
\bfseries rus & {\cellcolor[HTML]{81AED2}} \color[HTML]{000000} 0.8077 & {\cellcolor[HTML]{D5D5E8}} \color[HTML]{000000} 0.4952 & {\cellcolor[HTML]{FFF7FB}} \color[HTML]{000000} - \\
\bfseries spa & {\cellcolor[HTML]{8BB2D4}} \color[HTML]{000000} 0.7788 & {\cellcolor[HTML]{B7C5DF}} \color[HTML]{000000} 0.6256 & {\cellcolor[HTML]{E7E3F0}} \color[HTML]{000000} 0.3965 \\
\bfseries swa & {\cellcolor[HTML]{8EB3D5}} \color[HTML]{000000} 0.7658 & {\cellcolor[HTML]{E2DFEE}} \color[HTML]{000000} 0.4226 & {\cellcolor[HTML]{DCDAEB}} \color[HTML]{000000} 0.4570 \\
\bfseries tel & {\cellcolor[HTML]{67A4CC}} \color[HTML]{000000} 0.8818 & {\cellcolor[HTML]{F8F1F8}} \color[HTML]{000000} 0.2573 & {\cellcolor[HTML]{FEF6FA}} \color[HTML]{000000} 0.2143 \\
\bfseries tur & {\cellcolor[HTML]{83AFD3}} \color[HTML]{000000} 0.8008 & {\cellcolor[HTML]{C5CCE3}} \color[HTML]{000000} 0.5692 & {\cellcolor[HTML]{E4E1EF}} \color[HTML]{000000} 0.4125 \\
\bfseries urd & {\cellcolor[HTML]{8BB2D4}} \color[HTML]{000000} 0.7743 & {\cellcolor[HTML]{89B1D4}} \color[HTML]{000000} 0.7791 & {\cellcolor[HTML]{80AED2}} \color[HTML]{000000} 0.8108 \\
\bfseries zho & {\cellcolor[HTML]{589EC8}} \color[HTML]{000000} 0.9237 & {\cellcolor[HTML]{7DACD1}} \color[HTML]{000000} 0.8199 & {\cellcolor[HTML]{D6D6E9}} \color[HTML]{000000} 0.4912 \\
\hline
\end{tabular}
}
\caption{Per-language performance of the submitted system based on official results.}
\label{tab:performance}
\end{table}
\begin{table*}[!t]
\centering
\resizebox{\linewidth}{!}{
\begin{tabular}{r|c|c|m{1.1cm}|c|m{1.1cm}|c|c|c|m{1.5cm}|m{1.5cm}|m{1.5cm}|c}
\hline
& \textbf{Subtask 1} & \multicolumn{5}{c|}{\textbf{Subtask 2}} & \multicolumn{6}{c}{\textbf{Subtask 3}}\\
\bfseries Lang. & \bfseries polarization & \bfseries political & \bfseries racial/ ethnic & \bfseries religious & \bfseries gender/ sexual & \bfseries other & \bfseries stereotype & \bfseries vilification & \bfseries dehuman\-ization & \bfseries extreme language & \bfseries lack of empathy & \bfseries invalidation \\
\hline
\bfseries amh & {\cellcolor[HTML]{ADC1DD}} \color[HTML]{000000} 0.6619 & {\cellcolor[HTML]{94B6D7}} \color[HTML]{000000} 0.7447 & {\cellcolor[HTML]{93B5D6}} \color[HTML]{000000} 0.7533 & {\cellcolor[HTML]{9CB9D9}} \color[HTML]{000000} 0.7180 & {\cellcolor[HTML]{BDC8E1}} \color[HTML]{000000} 0.5987 & {\cellcolor[HTML]{A8BEDC}} \color[HTML]{000000} 0.6815 & {\cellcolor[HTML]{A8BEDC}} \color[HTML]{000000} 0.6805 & {\cellcolor[HTML]{99B8D8}} \color[HTML]{000000} 0.7282 & {\cellcolor[HTML]{A4BCDA}} \color[HTML]{000000} 0.6927 & {\cellcolor[HTML]{BDC8E1}} \color[HTML]{000000} 0.6003 & {\cellcolor[HTML]{DDDBEC}} \color[HTML]{000000} 0.4516 & {\cellcolor[HTML]{CACEE5}} \color[HTML]{000000} 0.5488 \\
\bfseries arb & {\cellcolor[HTML]{78ABD0}} \color[HTML]{000000} 0.8348 & {\cellcolor[HTML]{78ABD0}} \color[HTML]{000000} 0.8382 & {\cellcolor[HTML]{8BB2D4}} \color[HTML]{000000} 0.7787 & {\cellcolor[HTML]{73A9CF}} \color[HTML]{000000} 0.8542 & {\cellcolor[HTML]{91B5D6}} \color[HTML]{000000} 0.7577 & {\cellcolor[HTML]{A5BDDB}} \color[HTML]{000000} 0.6902 & {\cellcolor[HTML]{80AED2}} \color[HTML]{000000} 0.8096 & {\cellcolor[HTML]{79ABD0}} \color[HTML]{000000} 0.8322 & {\cellcolor[HTML]{9EBAD9}} \color[HTML]{000000} 0.7176 & {\cellcolor[HTML]{81AED2}} \color[HTML]{000000} 0.8084 & {\cellcolor[HTML]{BDC8E1}} \color[HTML]{000000} 0.6004 & {\cellcolor[HTML]{D2D3E7}} \color[HTML]{000000} 0.5111 \\
\bfseries ben & {\cellcolor[HTML]{76AAD0}} \color[HTML]{000000} 0.8415 & {\cellcolor[HTML]{80AED2}} \color[HTML]{000000} 0.8134 & {\cellcolor[HTML]{D5D5E8}} \color[HTML]{000000} 0.4972 & {\cellcolor[HTML]{B0C2DE}} \color[HTML]{000000} 0.6513 & {\cellcolor[HTML]{BDC8E1}} \color[HTML]{000000} 0.5987 & {\cellcolor[HTML]{BCC7E1}} \color[HTML]{000000} 0.6036 & {\cellcolor[HTML]{D7D6E9}} \color[HTML]{000000} 0.4847 & {\cellcolor[HTML]{9FBAD9}} \color[HTML]{000000} 0.7098 & {\cellcolor[HTML]{D6D6E9}} \color[HTML]{000000} 0.4899 & {\cellcolor[HTML]{C4CBE3}} \color[HTML]{000000} 0.5738 & {\cellcolor[HTML]{D5D5E8}} \color[HTML]{000000} 0.4951 & {\cellcolor[HTML]{D5D5E8}} \color[HTML]{000000} 0.4955 \\
\bfseries deu & {\cellcolor[HTML]{96B6D7}} \color[HTML]{000000} 0.7398 & {\cellcolor[HTML]{9CB9D9}} \color[HTML]{000000} 0.7222 & {\cellcolor[HTML]{91B5D6}} \color[HTML]{000000} 0.7552 & {\cellcolor[HTML]{81AED2}} \color[HTML]{000000} 0.8059 & {\cellcolor[HTML]{83AFD3}} \color[HTML]{000000} 0.8018 & {\cellcolor[HTML]{CACEE5}} \color[HTML]{000000} 0.5470 & {\cellcolor[HTML]{9EBAD9}} \color[HTML]{000000} 0.7138 & {\cellcolor[HTML]{A5BDDB}} \color[HTML]{000000} 0.6884 & {\cellcolor[HTML]{B7C5DF}} \color[HTML]{000000} 0.6239 & {\cellcolor[HTML]{A8BEDC}} \color[HTML]{000000} 0.6780 & {\cellcolor[HTML]{BCC7E1}} \color[HTML]{000000} 0.6039 & {\cellcolor[HTML]{DAD9EA}} \color[HTML]{000000} 0.4646 \\
\bfseries eng & {\cellcolor[HTML]{81AED2}} \color[HTML]{000000} 0.8058 & {\cellcolor[HTML]{83AFD3}} \color[HTML]{000000} 0.8014 & {\cellcolor[HTML]{97B7D7}} \color[HTML]{000000} 0.7353 & {\cellcolor[HTML]{93B5D6}} \color[HTML]{000000} 0.7535 & {\cellcolor[HTML]{A1BBDA}} \color[HTML]{000000} 0.7034 & {\cellcolor[HTML]{D2D2E7}} \color[HTML]{000000} 0.5194 & {\cellcolor[HTML]{A7BDDB}} \color[HTML]{000000} 0.6868 & {\cellcolor[HTML]{89B1D4}} \color[HTML]{000000} 0.7821 & {\cellcolor[HTML]{CDD0E5}} \color[HTML]{000000} 0.5391 & {\cellcolor[HTML]{91B5D6}} \color[HTML]{000000} 0.7560 & {\cellcolor[HTML]{C4CBE3}} \color[HTML]{000000} 0.5742 & {\cellcolor[HTML]{D7D6E9}} \color[HTML]{000000} 0.4894 \\
\bfseries fas & {\cellcolor[HTML]{8CB3D5}} \color[HTML]{000000} 0.7690 & {\cellcolor[HTML]{7DACD1}} \color[HTML]{000000} 0.8198 & {\cellcolor[HTML]{D6D6E9}} \color[HTML]{000000} 0.4937 & {\cellcolor[HTML]{75A9CF}} \color[HTML]{000000} 0.8482 & {\cellcolor[HTML]{9CB9D9}} \color[HTML]{000000} 0.7202 & {\cellcolor[HTML]{89B1D4}} \color[HTML]{000000} 0.7793 & {\cellcolor[HTML]{BFC9E1}} \color[HTML]{000000} 0.5919 & {\cellcolor[HTML]{A2BCDA}} \color[HTML]{000000} 0.7004 & {\cellcolor[HTML]{CCCFE5}} \color[HTML]{000000} 0.5445 & {\cellcolor[HTML]{B0C2DE}} \color[HTML]{000000} 0.6492 & {\cellcolor[HTML]{A9BFDC}} \color[HTML]{000000} 0.6763 & {\cellcolor[HTML]{D8D7E9}} \color[HTML]{000000} 0.4793 \\
\bfseries hau & {\cellcolor[HTML]{96B6D7}} \color[HTML]{000000} 0.7401 & {\cellcolor[HTML]{A8BEDC}} \color[HTML]{000000} 0.6783 & {\cellcolor[HTML]{B8C6E0}} \color[HTML]{000000} 0.6176 & {\cellcolor[HTML]{BCC7E1}} \color[HTML]{000000} 0.6031 & {\cellcolor[HTML]{D5D5E8}} \color[HTML]{000000} 0.4979 & {\cellcolor[HTML]{D5D5E8}} \color[HTML]{000000} 0.4991 & {\cellcolor[HTML]{D7D6E9}} \color[HTML]{000000} 0.4891 & {\cellcolor[HTML]{D5D5E8}} \color[HTML]{000000} 0.4966 & {\cellcolor[HTML]{D6D6E9}} \color[HTML]{000000} 0.4907 & {\cellcolor[HTML]{D6D6E9}} \color[HTML]{000000} 0.4920 & {\cellcolor[HTML]{D5D5E8}} \color[HTML]{000000} 0.4979 & {\cellcolor[HTML]{D5D5E8}} \color[HTML]{000000} 0.4994 \\
\bfseries hin & {\cellcolor[HTML]{84B0D3}} \color[HTML]{000000} 0.7974 & {\cellcolor[HTML]{83AFD3}} \color[HTML]{000000} 0.8019 & {\cellcolor[HTML]{65A3CB}} \color[HTML]{000000} 0.8896 & {\cellcolor[HTML]{589EC8}} \color[HTML]{000000} 0.9214 & {\cellcolor[HTML]{67A4CC}} \color[HTML]{000000} 0.8853 & {\cellcolor[HTML]{AFC1DD}} \color[HTML]{000000} 0.6536 & {\cellcolor[HTML]{9EBAD9}} \color[HTML]{000000} 0.7137 & {\cellcolor[HTML]{96B6D7}} \color[HTML]{000000} 0.7407 & {\cellcolor[HTML]{9EBAD9}} \color[HTML]{000000} 0.7130 & {\cellcolor[HTML]{ABBFDC}} \color[HTML]{000000} 0.6674 & {\cellcolor[HTML]{AFC1DD}} \color[HTML]{000000} 0.6554 & {\cellcolor[HTML]{94B6D7}} \color[HTML]{000000} 0.7480 \\
\bfseries ita & {\cellcolor[HTML]{99B8D8}} \color[HTML]{000000} 0.7303 & {\cellcolor[HTML]{E2DFEE}} \color[HTML]{000000} 0.4227 & {\cellcolor[HTML]{8BB2D4}} \color[HTML]{000000} 0.7741 & {\cellcolor[HTML]{78ABD0}} \color[HTML]{000000} 0.8374 & {\cellcolor[HTML]{BDC8E1}} \color[HTML]{000000} 0.6011 & {\cellcolor[HTML]{DBDAEB}} \color[HTML]{000000} 0.4617 & {\cellcolor[HTML]{FFF7FB}} \color[HTML]{000000} - & {\cellcolor[HTML]{FFF7FB}} \color[HTML]{000000} - & {\cellcolor[HTML]{FFF7FB}} \color[HTML]{000000} - & {\cellcolor[HTML]{FFF7FB}} \color[HTML]{000000} - & {\cellcolor[HTML]{FFF7FB}} \color[HTML]{000000} - & {\cellcolor[HTML]{FFF7FB}} \color[HTML]{000000} - \\
\bfseries khm & {\cellcolor[HTML]{B5C4DF}} \color[HTML]{000000} 0.6293 & {\cellcolor[HTML]{81AED2}} \color[HTML]{000000} 0.8052 & {\cellcolor[HTML]{9CB9D9}} \color[HTML]{000000} 0.7220 & {\cellcolor[HTML]{6DA6CD}} \color[HTML]{000000} 0.8673 & {\cellcolor[HTML]{9CB9D9}} \color[HTML]{000000} 0.7207 & {\cellcolor[HTML]{88B1D4}} \color[HTML]{000000} 0.7873 & {\cellcolor[HTML]{9AB8D8}} \color[HTML]{000000} 0.7275 & {\cellcolor[HTML]{CACEE5}} \color[HTML]{000000} 0.5464 & {\cellcolor[HTML]{D1D2E6}} \color[HTML]{000000} 0.5233 & {\cellcolor[HTML]{D6D6E9}} \color[HTML]{000000} 0.4942 & {\cellcolor[HTML]{A4BCDA}} \color[HTML]{000000} 0.6960 & {\cellcolor[HTML]{D2D2E7}} \color[HTML]{000000} 0.5168 \\
\bfseries mya & {\cellcolor[HTML]{69A5CC}} \color[HTML]{000000} 0.8788 & {\cellcolor[HTML]{63A2CB}} \color[HTML]{000000} 0.8955 & {\cellcolor[HTML]{96B6D7}} \color[HTML]{000000} 0.7417 & {\cellcolor[HTML]{81AED2}} \color[HTML]{000000} 0.8071 & {\cellcolor[HTML]{86B0D3}} \color[HTML]{000000} 0.7927 & {\cellcolor[HTML]{76AAD0}} \color[HTML]{000000} 0.8399 & {\cellcolor[HTML]{FFF7FB}} \color[HTML]{000000} - & {\cellcolor[HTML]{FFF7FB}} \color[HTML]{000000} - & {\cellcolor[HTML]{FFF7FB}} \color[HTML]{000000} - & {\cellcolor[HTML]{FFF7FB}} \color[HTML]{000000} - & {\cellcolor[HTML]{FFF7FB}} \color[HTML]{000000} - & {\cellcolor[HTML]{FFF7FB}} \color[HTML]{000000} - \\
\bfseries nep & {\cellcolor[HTML]{63A2CB}} \color[HTML]{000000} 0.8915 & {\cellcolor[HTML]{76AAD0}} \color[HTML]{000000} 0.8438 & {\cellcolor[HTML]{569DC8}} \color[HTML]{000000} 0.9281 & {\cellcolor[HTML]{529BC7}} \color[HTML]{000000} 0.9369 & {\cellcolor[HTML]{5C9FC9}} \color[HTML]{000000} 0.9132 & {\cellcolor[HTML]{7EADD1}} \color[HTML]{000000} 0.8187 & {\cellcolor[HTML]{6FA7CE}} \color[HTML]{000000} 0.8650 & {\cellcolor[HTML]{84B0D3}} \color[HTML]{000000} 0.7952 & {\cellcolor[HTML]{91B5D6}} \color[HTML]{000000} 0.7537 & {\cellcolor[HTML]{7DACD1}} \color[HTML]{000000} 0.8244 & {\cellcolor[HTML]{ACC0DD}} \color[HTML]{000000} 0.6643 & {\cellcolor[HTML]{C8CDE4}} \color[HTML]{000000} 0.5589 \\
\bfseries ori & {\cellcolor[HTML]{83AFD3}} \color[HTML]{000000} 0.8013 & {\cellcolor[HTML]{89B1D4}} \color[HTML]{000000} 0.7828 & {\cellcolor[HTML]{99B8D8}} \color[HTML]{000000} 0.7305 & {\cellcolor[HTML]{88B1D4}} \color[HTML]{000000} 0.7847 & {\cellcolor[HTML]{B7C5DF}} \color[HTML]{000000} 0.6254 & {\cellcolor[HTML]{ADC1DD}} \color[HTML]{000000} 0.6590 & {\cellcolor[HTML]{C4CBE3}} \color[HTML]{000000} 0.5720 & {\cellcolor[HTML]{B1C2DE}} \color[HTML]{000000} 0.6428 & {\cellcolor[HTML]{D5D5E8}} \color[HTML]{000000} 0.4984 & {\cellcolor[HTML]{C5CCE3}} \color[HTML]{000000} 0.5680 & {\cellcolor[HTML]{D5D5E8}} \color[HTML]{000000} 0.4960 & {\cellcolor[HTML]{D6D6E9}} \color[HTML]{000000} 0.4914 \\
\bfseries pan & {\cellcolor[HTML]{8CB3D5}} \color[HTML]{000000} 0.7736 & {\cellcolor[HTML]{9FBAD9}} \color[HTML]{000000} 0.7096 & {\cellcolor[HTML]{BFC9E1}} \color[HTML]{000000} 0.5914 & {\cellcolor[HTML]{A9BFDC}} \color[HTML]{000000} 0.6745 & {\cellcolor[HTML]{A5BDDB}} \color[HTML]{000000} 0.6923 & {\cellcolor[HTML]{C8CDE4}} \color[HTML]{000000} 0.5558 & {\cellcolor[HTML]{B3C3DE}} \color[HTML]{000000} 0.6384 & {\cellcolor[HTML]{91B5D6}} \color[HTML]{000000} 0.7562 & {\cellcolor[HTML]{9CB9D9}} \color[HTML]{000000} 0.7214 & {\cellcolor[HTML]{A7BDDB}} \color[HTML]{000000} 0.6834 & {\cellcolor[HTML]{CCCFE5}} \color[HTML]{000000} 0.5423 & {\cellcolor[HTML]{AFC1DD}} \color[HTML]{000000} 0.6547 \\
\bfseries pol & {\cellcolor[HTML]{7EADD1}} \color[HTML]{000000} 0.8158 & {\cellcolor[HTML]{79ABD0}} \color[HTML]{000000} 0.8314 & {\cellcolor[HTML]{97B7D7}} \color[HTML]{000000} 0.7368 & {\cellcolor[HTML]{73A9CF}} \color[HTML]{000000} 0.8523 & {\cellcolor[HTML]{8BB2D4}} \color[HTML]{000000} 0.7761 & {\cellcolor[HTML]{CDD0E5}} \color[HTML]{000000} 0.5392 & {\cellcolor[HTML]{FFF7FB}} \color[HTML]{000000} - & {\cellcolor[HTML]{FFF7FB}} \color[HTML]{000000} - & {\cellcolor[HTML]{FFF7FB}} \color[HTML]{000000} - & {\cellcolor[HTML]{FFF7FB}} \color[HTML]{000000} - & {\cellcolor[HTML]{FFF7FB}} \color[HTML]{000000} - & {\cellcolor[HTML]{FFF7FB}} \color[HTML]{000000} - \\
\bfseries rus & {\cellcolor[HTML]{81AED2}} \color[HTML]{000000} 0.8077 & {\cellcolor[HTML]{7DACD1}} \color[HTML]{000000} 0.8231 & {\cellcolor[HTML]{94B6D7}} \color[HTML]{000000} 0.7451 & {\cellcolor[HTML]{7DACD1}} \color[HTML]{000000} 0.8207 & {\cellcolor[HTML]{96B6D7}} \color[HTML]{000000} 0.7389 & {\cellcolor[HTML]{CACEE5}} \color[HTML]{000000} 0.5469 & {\cellcolor[HTML]{FFF7FB}} \color[HTML]{000000} - & {\cellcolor[HTML]{FFF7FB}} \color[HTML]{000000} - & {\cellcolor[HTML]{FFF7FB}} \color[HTML]{000000} - & {\cellcolor[HTML]{FFF7FB}} \color[HTML]{000000} - & {\cellcolor[HTML]{FFF7FB}} \color[HTML]{000000} - & {\cellcolor[HTML]{FFF7FB}} \color[HTML]{000000} - \\
\bfseries spa & {\cellcolor[HTML]{8BB2D4}} \color[HTML]{000000} 0.7788 & {\cellcolor[HTML]{81AED2}} \color[HTML]{000000} 0.8069 & {\cellcolor[HTML]{A2BCDA}} \color[HTML]{000000} 0.6994 & {\cellcolor[HTML]{8BB2D4}} \color[HTML]{000000} 0.7785 & {\cellcolor[HTML]{67A4CC}} \color[HTML]{000000} 0.8831 & {\cellcolor[HTML]{99B8D8}} \color[HTML]{000000} 0.7310 & {\cellcolor[HTML]{A9BFDC}} \color[HTML]{000000} 0.6754 & {\cellcolor[HTML]{93B5D6}} \color[HTML]{000000} 0.7521 & {\cellcolor[HTML]{D9D8EA}} \color[HTML]{000000} 0.4766 & {\cellcolor[HTML]{9CB9D9}} \color[HTML]{000000} 0.7227 & {\cellcolor[HTML]{C6CCE3}} \color[HTML]{000000} 0.5639 & {\cellcolor[HTML]{B1C2DE}} \color[HTML]{000000} 0.6442 \\
\bfseries swa & {\cellcolor[HTML]{8EB3D5}} \color[HTML]{000000} 0.7658 & {\cellcolor[HTML]{A4BCDA}} \color[HTML]{000000} 0.6926 & {\cellcolor[HTML]{78ABD0}} \color[HTML]{000000} 0.8357 & {\cellcolor[HTML]{60A1CA}} \color[HTML]{000000} 0.9040 & {\cellcolor[HTML]{D3D4E7}} \color[HTML]{000000} 0.5085 & {\cellcolor[HTML]{CED0E6}} \color[HTML]{000000} 0.5304 & {\cellcolor[HTML]{91B5D6}} \color[HTML]{000000} 0.7573 & {\cellcolor[HTML]{96B6D7}} \color[HTML]{000000} 0.7393 & {\cellcolor[HTML]{DAD9EA}} \color[HTML]{000000} 0.4658 & {\cellcolor[HTML]{B1C2DE}} \color[HTML]{000000} 0.6425 & {\cellcolor[HTML]{A2BCDA}} \color[HTML]{000000} 0.6995 & {\cellcolor[HTML]{BBC7E0}} \color[HTML]{000000} 0.6073 \\
\bfseries tel & {\cellcolor[HTML]{67A4CC}} \color[HTML]{000000} 0.8818 & {\cellcolor[HTML]{C8CDE4}} \color[HTML]{000000} 0.5560 & {\cellcolor[HTML]{B9C6E0}} \color[HTML]{000000} 0.6134 & {\cellcolor[HTML]{C4CBE3}} \color[HTML]{000000} 0.5720 & {\cellcolor[HTML]{B8C6E0}} \color[HTML]{000000} 0.6190 & {\cellcolor[HTML]{CCCFE5}} \color[HTML]{000000} 0.5444 & {\cellcolor[HTML]{D9D8EA}} \color[HTML]{000000} 0.4702 & {\cellcolor[HTML]{B7C5DF}} \color[HTML]{000000} 0.6228 & {\cellcolor[HTML]{D6D6E9}} \color[HTML]{000000} 0.4938 & {\cellcolor[HTML]{D7D6E9}} \color[HTML]{000000} 0.4853 & {\cellcolor[HTML]{A1BBDA}} \color[HTML]{000000} 0.7034 & {\cellcolor[HTML]{C2CBE2}} \color[HTML]{000000} 0.5792 \\
\bfseries tur & {\cellcolor[HTML]{83AFD3}} \color[HTML]{000000} 0.8008 & {\cellcolor[HTML]{81AED2}} \color[HTML]{000000} 0.8092 & {\cellcolor[HTML]{86B0D3}} \color[HTML]{000000} 0.7910 & {\cellcolor[HTML]{73A9CF}} \color[HTML]{000000} 0.8515 & {\cellcolor[HTML]{88B1D4}} \color[HTML]{000000} 0.7867 & {\cellcolor[HTML]{D1D2E6}} \color[HTML]{000000} 0.5221 & {\cellcolor[HTML]{8EB3D5}} \color[HTML]{000000} 0.7673 & {\cellcolor[HTML]{91B5D6}} \color[HTML]{000000} 0.7570 & {\cellcolor[HTML]{C0C9E2}} \color[HTML]{000000} 0.5862 & {\cellcolor[HTML]{80AED2}} \color[HTML]{000000} 0.8122 & {\cellcolor[HTML]{D9D8EA}} \color[HTML]{000000} 0.4747 & {\cellcolor[HTML]{D7D6E9}} \color[HTML]{000000} 0.4861 \\
\bfseries urd & {\cellcolor[HTML]{8BB2D4}} \color[HTML]{000000} 0.7743 & {\cellcolor[HTML]{89B1D4}} \color[HTML]{000000} 0.7806 & {\cellcolor[HTML]{9CB9D9}} \color[HTML]{000000} 0.7192 & {\cellcolor[HTML]{97B7D7}} \color[HTML]{000000} 0.7379 & {\cellcolor[HTML]{9EBAD9}} \color[HTML]{000000} 0.7168 & {\cellcolor[HTML]{9FBAD9}} \color[HTML]{000000} 0.7103 & {\cellcolor[HTML]{93B5D6}} \color[HTML]{000000} 0.7527 & {\cellcolor[HTML]{8CB3D5}} \color[HTML]{000000} 0.7701 & {\cellcolor[HTML]{9AB8D8}} \color[HTML]{000000} 0.7276 & {\cellcolor[HTML]{8FB4D6}} \color[HTML]{000000} 0.7634 & {\cellcolor[HTML]{99B8D8}} \color[HTML]{000000} 0.7309 & {\cellcolor[HTML]{9AB8D8}} \color[HTML]{000000} 0.7248 \\
\bfseries zho & {\cellcolor[HTML]{589EC8}} \color[HTML]{000000} 0.9237 & {\cellcolor[HTML]{71A8CE}} \color[HTML]{000000} 0.8571 & {\cellcolor[HTML]{589EC8}} \color[HTML]{000000} 0.9212 & {\cellcolor[HTML]{4897C4}} \color[HTML]{000000} 0.9651 & {\cellcolor[HTML]{589EC8}} \color[HTML]{000000} 0.9245 & {\cellcolor[HTML]{7BACD1}} \color[HTML]{000000} 0.8294 & {\cellcolor[HTML]{78ABD0}} \color[HTML]{000000} 0.8393 & {\cellcolor[HTML]{6DA6CD}} \color[HTML]{000000} 0.8696 & {\cellcolor[HTML]{84B0D3}} \color[HTML]{000000} 0.7958 & {\cellcolor[HTML]{A2BCDA}} \color[HTML]{000000} 0.6997 & {\cellcolor[HTML]{CACEE5}} \color[HTML]{000000} 0.5491 & {\cellcolor[HTML]{BFC9E1}} \color[HTML]{000000} 0.5937 \\
\hline
\end{tabular}
}
\caption{Per-language per-feature (label) performance of the submitted system based on the released test-set gold labels.}
\label{tab:performance_labels}
\end{table*}
\begin{table}[!t]
\centering
\resizebox{0.95\linewidth}{!}{
\begin{tabular}{r|c|c|c}
\hline
 & \multicolumn{3}{c}{\textbf{Macro F1}}\\
\bfseries Language & \bfseries Subtask 1 & \bfseries Subtask 2 & \bfseries Subtask 3 \\
\hline
\bfseries amh & {\cellcolor[HTML]{FFF8B4}} \color[HTML]{000000} -0.0532 & {\cellcolor[HTML]{E9F6A1}} \color[HTML]{000000} 0.1400 & {\cellcolor[HTML]{FFFDBC}} \color[HTML]{000000} -0.0123 \\
\bfseries arb & {\cellcolor[HTML]{F8FCB6}} \color[HTML]{000000} 0.0391 & {\cellcolor[HTML]{E8F59F}} \color[HTML]{000000} 0.1424 & {\cellcolor[HTML]{EBF7A3}} \color[HTML]{000000} 0.1255 \\
\bfseries ben & {\cellcolor[HTML]{FFFDBC}} \color[HTML]{000000} -0.0113 & {\cellcolor[HTML]{FDFEBC}} \color[HTML]{000000} 0.0163 & {\cellcolor[HTML]{F8FCB6}} \color[HTML]{000000} 0.0404 \\
\bfseries deu & {\cellcolor[HTML]{F4FAB0}} \color[HTML]{000000} 0.0684 & {\cellcolor[HTML]{E9F6A1}} \color[HTML]{000000} 0.1321 & {\cellcolor[HTML]{F7FCB4}} \color[HTML]{000000} 0.0559 \\
\bfseries eng & {\cellcolor[HTML]{FBFDBA}} \color[HTML]{000000} 0.0256 & {\cellcolor[HTML]{ECF7A6}} \color[HTML]{000000} 0.1186 & {\cellcolor[HTML]{FFFAB6}} \color[HTML]{000000} -0.0403 \\
\bfseries fas & {\cellcolor[HTML]{FFF6B0}} \color[HTML]{000000} -0.0734 & {\cellcolor[HTML]{F5FBB2}} \color[HTML]{000000} 0.0624 & {\cellcolor[HTML]{ECF7A6}} \color[HTML]{000000} 0.1204 \\
\bfseries hau & {\cellcolor[HTML]{FFFBB8}} \color[HTML]{000000} -0.0352 & {\cellcolor[HTML]{FFFBB8}} \color[HTML]{000000} -0.0349 & {\cellcolor[HTML]{F16640}} \color[HTML]{F1F1F1} -0.7456 \\
\bfseries hin & {\cellcolor[HTML]{F5FBB2}} \color[HTML]{000000} 0.0595 & {\cellcolor[HTML]{FFFBB8}} \color[HTML]{000000} -0.0338 & {\cellcolor[HTML]{9DD569}} \color[HTML]{000000} 0.5105 \\
\bfseries ita & {\cellcolor[HTML]{F7FCB4}} \color[HTML]{000000} 0.0530 & {\cellcolor[HTML]{FFF6B0}} \color[HTML]{000000} -0.0740 & {\cellcolor[HTML]{FEFFBE}} \color[HTML]{000000} 0.0000 \\
\bfseries khm & {\cellcolor[HTML]{FFFBB8}} \color[HTML]{000000} -0.0299 & {\cellcolor[HTML]{FEFFBE}} \color[HTML]{000000} 0.0055 & {\cellcolor[HTML]{FDC776}} \color[HTML]{000000} -0.3613 \\
\bfseries mya & {\cellcolor[HTML]{F5FBB2}} \color[HTML]{000000} 0.0578 & {\cellcolor[HTML]{DDF191}} \color[HTML]{000000} 0.2063 & {\cellcolor[HTML]{FEFFBE}} \color[HTML]{000000} 0.0000 \\
\bfseries nep & {\cellcolor[HTML]{FDFEBC}} \color[HTML]{000000} 0.0117 & {\cellcolor[HTML]{F2FAAE}} \color[HTML]{000000} 0.0807 & {\cellcolor[HTML]{AFDD70}} \color[HTML]{000000} 0.4355 \\
\bfseries ori & {\cellcolor[HTML]{FBFDBA}} \color[HTML]{000000} 0.0248 & {\cellcolor[HTML]{FFF2AA}} \color[HTML]{000000} -0.0972 & {\cellcolor[HTML]{FEDC88}} \color[HTML]{000000} -0.2614 \\
\bfseries pan & {\cellcolor[HTML]{FFFDBC}} \color[HTML]{000000} -0.0162 & {\cellcolor[HTML]{FEFFBE}} \color[HTML]{000000} 0.0084 & {\cellcolor[HTML]{FAFDB8}} \color[HTML]{000000} 0.0372 \\
\bfseries pol & {\cellcolor[HTML]{F1F9AC}} \color[HTML]{000000} 0.0917 & {\cellcolor[HTML]{F1F9AC}} \color[HTML]{000000} 0.0859 & {\cellcolor[HTML]{FEFFBE}} \color[HTML]{000000} 0.0000 \\
\bfseries rus & {\cellcolor[HTML]{F5FBB2}} \color[HTML]{000000} 0.0620 & {\cellcolor[HTML]{FFF2AA}} \color[HTML]{000000} -0.0952 & {\cellcolor[HTML]{FEFFBE}} \color[HTML]{000000} 0.0000 \\
\bfseries spa & {\cellcolor[HTML]{F7FCB4}} \color[HTML]{000000} 0.0522 & {\cellcolor[HTML]{FAFDB8}} \color[HTML]{000000} 0.0321 & {\cellcolor[HTML]{FFF1A8}} \color[HTML]{000000} -0.1123 \\
\bfseries swa & {\cellcolor[HTML]{FEFFBE}} \color[HTML]{000000} 0.0087 & {\cellcolor[HTML]{FFFCBA}} \color[HTML]{000000} -0.0191 & {\cellcolor[HTML]{D9EF8B}} \color[HTML]{000000} 0.2365 \\
\bfseries tel & {\cellcolor[HTML]{D9EF8B}} \color[HTML]{000000} 0.2378 & {\cellcolor[HTML]{FFF7B2}} \color[HTML]{000000} -0.0572 & {\cellcolor[HTML]{FDB163}} \color[HTML]{000000} -0.4595 \\
\bfseries tur & {\cellcolor[HTML]{EEF8A8}} \color[HTML]{000000} 0.1051 & {\cellcolor[HTML]{EFF8AA}} \color[HTML]{000000} 0.0984 & {\cellcolor[HTML]{FDC776}} \color[HTML]{000000} -0.3568 \\
\bfseries urd & {\cellcolor[HTML]{FFFDBC}} \color[HTML]{000000} -0.0147 & {\cellcolor[HTML]{F4FAB0}} \color[HTML]{000000} 0.0664 & {\cellcolor[HTML]{D1EC86}} \color[HTML]{000000} 0.2792 \\
\bfseries zho & {\cellcolor[HTML]{F7FCB4}} \color[HTML]{000000} 0.0546 & {\cellcolor[HTML]{E6F59D}} \color[HTML]{000000} 0.1502 & {\cellcolor[HTML]{A2D76A}} \color[HTML]{000000} 0.4912 \\
\bfseries Average & {\cellcolor[HTML]{FAFDB8}} \color[HTML]{000000} 0.0326 & {\cellcolor[HTML]{F8FCB6}} \color[HTML]{000000} 0.0425 & {\cellcolor[HTML]{FFFEBE}} \color[HTML]{000000} -0.0008 \\
\hline
\end{tabular}
}
\caption{Per-language performance comparison to the official baseline.}
\label{tab:baseline}
\end{table}

Based on the lowest achieved macro F1 value in subtask 1 (0.63 in case of Khmer), we can see that the submitted detection system can distinguish the two classes (very well above chance). It performs much worse in the other two subtasks, where multi-label nature makes the detection challenging. Only in 6 cases, it was able to achieve above 0.7 macro F1 value.

In comparison to the baseline included in the official results, the submitted system performs better in about two thirds of cases (see Table~\ref{tab:baseline}). On average, it achieved by 3\% in subtask 1 and by 4\% in subtask 2 higher Macro F1 score. The performance in subtask 3 was about the same on average, highly influenced by F1 score of 0.0 in Hausa. The highest increase is in Hindi (by 51\%).

\begin{table}[!t]
\centering
\resizebox{0.95\linewidth}{!}{
\begin{tabular}{r|c|c|c}
\hline
 & \multicolumn{3}{c}{\textbf{Percentile}}\\
\bfseries Language & \bfseries Subtask 1 & \bfseries Subtask 2 & \bfseries Subtask 3 \\
\hline
\bfseries amh & {\cellcolor[HTML]{F0F9ED}} \color[HTML]{000000} 7.1 & {\cellcolor[HTML]{C2E7BB}} \color[HTML]{000000} 40.0 & {\cellcolor[HTML]{DAF0D4}} \color[HTML]{000000} 26.3 \\
\bfseries arb & {\cellcolor[HTML]{55B567}} \color[HTML]{000000} 86.7 & {\cellcolor[HTML]{6DC072}} \color[HTML]{000000} 77.8 & {\cellcolor[HTML]{E8F6E4}} \color[HTML]{000000} 15.8 \\
\bfseries ben & {\cellcolor[HTML]{5DB96B}} \color[HTML]{000000} 83.7 & {\cellcolor[HTML]{B5E1AE}} \color[HTML]{000000} 46.7 & {\cellcolor[HTML]{DDF2D8}} \color[HTML]{000000} 23.8 \\
\bfseries deu & {\cellcolor[HTML]{4EB264}} \color[HTML]{000000} 88.9 & {\cellcolor[HTML]{AEDEA7}} \color[HTML]{000000} 50.0 & {\cellcolor[HTML]{E2F4DD}} \color[HTML]{000000} 21.1 \\
\bfseries eng & {\cellcolor[HTML]{6BC072}} \color[HTML]{000000} 78.3 & {\cellcolor[HTML]{CEECC8}} \color[HTML]{000000} 33.3 & {\cellcolor[HTML]{EFF9EC}} \color[HTML]{000000} 8.3 \\
\bfseries fas & {\cellcolor[HTML]{ECF8E8}} \color[HTML]{000000} 11.4 & {\cellcolor[HTML]{E9F7E5}} \color[HTML]{000000} 14.8 & {\cellcolor[HTML]{EDF8EA}} \color[HTML]{000000} 10.5 \\
\bfseries hau & {\cellcolor[HTML]{D9F0D3}} \color[HTML]{000000} 26.7 & {\cellcolor[HTML]{D6EFD0}} \color[HTML]{000000} 28.6 & {\cellcolor[HTML]{F7FCF5}} \color[HTML]{000000} 0.0 \\
\bfseries hin & {\cellcolor[HTML]{A8DCA2}} \color[HTML]{000000} 53.2 & {\cellcolor[HTML]{B5E1AE}} \color[HTML]{000000} 46.7 & {\cellcolor[HTML]{7DC87E}} \color[HTML]{000000} 71.4 \\
\bfseries ita & {\cellcolor[HTML]{3BA458}} \color[HTML]{000000} 97.7 & {\cellcolor[HTML]{9ED798}} \color[HTML]{000000} 57.7 & {\cellcolor[HTML]{F7FCF5}} \color[HTML]{000000} - \\
\bfseries khm & {\cellcolor[HTML]{D7EFD1}} \color[HTML]{000000} 27.9 & {\cellcolor[HTML]{B2E0AC}} \color[HTML]{000000} 48.0 & {\cellcolor[HTML]{DAF0D4}} \color[HTML]{000000} 26.3 \\
\bfseries mya & {\cellcolor[HTML]{5DB96B}} \color[HTML]{000000} 83.3 & {\cellcolor[HTML]{AADDA4}} \color[HTML]{000000} 52.0 & {\cellcolor[HTML]{F7FCF5}} \color[HTML]{000000} - \\
\bfseries nep & {\cellcolor[HTML]{D8F0D2}} \color[HTML]{000000} 27.3 & {\cellcolor[HTML]{4DB163}} \color[HTML]{000000} 89.3 & {\cellcolor[HTML]{DBF1D6}} \color[HTML]{000000} 25.0 \\
\bfseries ori & {\cellcolor[HTML]{60BA6C}} \color[HTML]{000000} 82.2 & {\cellcolor[HTML]{C1E6BA}} \color[HTML]{000000} 40.7 & {\cellcolor[HTML]{E5F5E0}} \color[HTML]{000000} 19.0 \\
\bfseries pan & {\cellcolor[HTML]{A0D99B}} \color[HTML]{000000} 56.8 & {\cellcolor[HTML]{EBF7E7}} \color[HTML]{000000} 12.5 & {\cellcolor[HTML]{B8E3B2}} \color[HTML]{000000} 45.0 \\
\bfseries pol & {\cellcolor[HTML]{56B567}} \color[HTML]{000000} 86.0 & {\cellcolor[HTML]{99D595}} \color[HTML]{000000} 59.3 & {\cellcolor[HTML]{F7FCF5}} \color[HTML]{000000} - \\
\bfseries rus & {\cellcolor[HTML]{56B567}} \color[HTML]{000000} 86.0 & {\cellcolor[HTML]{C8E9C1}} \color[HTML]{000000} 37.0 & {\cellcolor[HTML]{F7FCF5}} \color[HTML]{000000} - \\
\bfseries spa & {\cellcolor[HTML]{80CA80}} \color[HTML]{000000} 70.0 & {\cellcolor[HTML]{9BD696}} \color[HTML]{000000} 58.6 & {\cellcolor[HTML]{EEF8EA}} \color[HTML]{000000} 9.5 \\
\bfseries swa & {\cellcolor[HTML]{E5F5E1}} \color[HTML]{000000} 18.2 & {\cellcolor[HTML]{CEECC8}} \color[HTML]{000000} 33.3 & {\cellcolor[HTML]{EDF8EA}} \color[HTML]{000000} 10.0 \\
\bfseries tel & {\cellcolor[HTML]{7DC87E}} \color[HTML]{000000} 71.1 & {\cellcolor[HTML]{E5F5E1}} \color[HTML]{000000} 18.5 & {\cellcolor[HTML]{DDF2D8}} \color[HTML]{000000} 23.8 \\
\bfseries tur & {\cellcolor[HTML]{56B567}} \color[HTML]{000000} 85.7 & {\cellcolor[HTML]{A7DBA0}} \color[HTML]{000000} 53.8 & {\cellcolor[HTML]{F2FAF0}} \color[HTML]{000000} 5.3 \\
\bfseries urd & {\cellcolor[HTML]{AEDEA7}} \color[HTML]{000000} 50.0 & {\cellcolor[HTML]{5DB96B}} \color[HTML]{000000} 83.3 & {\cellcolor[HTML]{63BC6E}} \color[HTML]{000000} 81.0 \\
\bfseries zho & {\cellcolor[HTML]{4DB163}} \color[HTML]{000000} 89.1 & {\cellcolor[HTML]{66BD6F}} \color[HTML]{000000} 80.0 & {\cellcolor[HTML]{EAF7E6}} \color[HTML]{000000} 13.6 \\
\hline
\end{tabular}
}
\caption{Per-language rank percentile of the submitted system based on unofficial results. Higher percentile is better in comparison to others.}
\label{tab:rank}
\end{table}
\begin{table*}[!t]
\centering
\resizebox{\linewidth}{!}{
\begin{tabular}{r|c|c|m{1.1cm}|c|m{1.1cm}|c|c|c|m{1.5cm}|m{1.5cm}|m{1.5cm}|c}
\hline
& \textbf{Subtask 1} & \multicolumn{5}{c|}{\textbf{Subtask 2}} & \multicolumn{6}{c}{\textbf{Subtask 3}}\\
\bfseries Lang. & \bfseries polarization & \bfseries political & \bfseries racial/ ethnic & \bfseries religious & \bfseries gender/ sexual & \bfseries other & \bfseries stereotype & \bfseries vilification & \bfseries dehuman\-ization & \bfseries extreme language & \bfseries lack of empathy & \bfseries invalidation \\
\hline
\bfseries amh & {\cellcolor[HTML]{A8BEDC}} \color[HTML]{000000} 0.6804 & {\cellcolor[HTML]{A9BFDC}} \color[HTML]{000000} 0.6730 & {\cellcolor[HTML]{A9BFDC}} \color[HTML]{000000} 0.6736 & {\cellcolor[HTML]{B9C6E0}} \color[HTML]{000000} 0.6119 & {\cellcolor[HTML]{DDDBEC}} \color[HTML]{000000} 0.4525 & {\cellcolor[HTML]{B1C2DE}} \color[HTML]{000000} 0.6442 & {\cellcolor[HTML]{B5C4DF}} \color[HTML]{000000} 0.6278 & {\cellcolor[HTML]{9EBAD9}} \color[HTML]{000000} 0.7156 & {\cellcolor[HTML]{BDC8E1}} \color[HTML]{000000} 0.5981 & {\cellcolor[HTML]{BFC9E1}} \color[HTML]{000000} 0.5955 & {\cellcolor[HTML]{D2D2E7}} \color[HTML]{000000} 0.5175 & {\cellcolor[HTML]{DAD9EA}} \color[HTML]{000000} 0.4666 \\
\bfseries arb & {\cellcolor[HTML]{A5BDDB}} \color[HTML]{000000} 0.6908 & {\cellcolor[HTML]{A1BBDA}} \color[HTML]{000000} 0.7041 & {\cellcolor[HTML]{B4C4DF}} \color[HTML]{000000} 0.6336 & {\cellcolor[HTML]{B4C4DF}} \color[HTML]{000000} 0.6365 & {\cellcolor[HTML]{94B6D7}} \color[HTML]{000000} 0.7463 & {\cellcolor[HTML]{9AB8D8}} \color[HTML]{000000} 0.7232 & {\cellcolor[HTML]{9FBAD9}} \color[HTML]{000000} 0.7105 & {\cellcolor[HTML]{99B8D8}} \color[HTML]{000000} 0.7306 & {\cellcolor[HTML]{A9BFDC}} \color[HTML]{000000} 0.6755 & {\cellcolor[HTML]{A5BDDB}} \color[HTML]{000000} 0.6903 & {\cellcolor[HTML]{A7BDDB}} \color[HTML]{000000} 0.6845 & {\cellcolor[HTML]{B4C4DF}} \color[HTML]{000000} 0.6336 \\
\bfseries ben & {\cellcolor[HTML]{8FB4D6}} \color[HTML]{000000} 0.7616 & {\cellcolor[HTML]{94B6D7}} \color[HTML]{000000} 0.7437 & {\cellcolor[HTML]{EEE9F3}} \color[HTML]{000000} 0.3429 & {\cellcolor[HTML]{A2BCDA}} \color[HTML]{000000} 0.6989 & {\cellcolor[HTML]{B0C2DE}} \color[HTML]{000000} 0.6501 & {\cellcolor[HTML]{9EBAD9}} \color[HTML]{000000} 0.7159 & {\cellcolor[HTML]{B7C5DF}} \color[HTML]{000000} 0.6252 & {\cellcolor[HTML]{94B6D7}} \color[HTML]{000000} 0.7445 & {\cellcolor[HTML]{ABBFDC}} \color[HTML]{000000} 0.6716 & {\cellcolor[HTML]{97B7D7}} \color[HTML]{000000} 0.7356 & {\cellcolor[HTML]{AFC1DD}} \color[HTML]{000000} 0.6544 & {\cellcolor[HTML]{9EBAD9}} \color[HTML]{000000} 0.7139 \\
\bfseries deu & {\cellcolor[HTML]{A9BFDC}} \color[HTML]{000000} 0.6757 & {\cellcolor[HTML]{ADC1DD}} \color[HTML]{000000} 0.6610 & {\cellcolor[HTML]{B9C6E0}} \color[HTML]{000000} 0.6127 & {\cellcolor[HTML]{A5BDDB}} \color[HTML]{000000} 0.6877 & {\cellcolor[HTML]{99B8D8}} \color[HTML]{000000} 0.7300 & {\cellcolor[HTML]{B9C6E0}} \color[HTML]{000000} 0.6135 & {\cellcolor[HTML]{ADC1DD}} \color[HTML]{000000} 0.6608 & {\cellcolor[HTML]{AFC1DD}} \color[HTML]{000000} 0.6520 & {\cellcolor[HTML]{ADC1DD}} \color[HTML]{000000} 0.6586 & {\cellcolor[HTML]{ADC1DD}} \color[HTML]{000000} 0.6593 & {\cellcolor[HTML]{C4CBE3}} \color[HTML]{000000} 0.5729 & {\cellcolor[HTML]{C8CDE4}} \color[HTML]{000000} 0.5564 \\
\bfseries eng & {\cellcolor[HTML]{96B6D7}} \color[HTML]{000000} 0.7411 & {\cellcolor[HTML]{9FBAD9}} \color[HTML]{000000} 0.7127 & {\cellcolor[HTML]{A5BDDB}} \color[HTML]{000000} 0.6892 & {\cellcolor[HTML]{94B6D7}} \color[HTML]{000000} 0.7476 & {\cellcolor[HTML]{ABBFDC}} \color[HTML]{000000} 0.6712 & {\cellcolor[HTML]{B4C4DF}} \color[HTML]{000000} 0.6345 & {\cellcolor[HTML]{94B6D7}} \color[HTML]{000000} 0.7446 & {\cellcolor[HTML]{A1BBDA}} \color[HTML]{000000} 0.7062 & {\cellcolor[HTML]{AFC1DD}} \color[HTML]{000000} 0.6559 & {\cellcolor[HTML]{A7BDDB}} \color[HTML]{000000} 0.6829 & {\cellcolor[HTML]{B1C2DE}} \color[HTML]{000000} 0.6463 & {\cellcolor[HTML]{ACC0DD}} \color[HTML]{000000} 0.6667 \\
\bfseries fas & {\cellcolor[HTML]{A8BEDC}} \color[HTML]{000000} 0.6797 & {\cellcolor[HTML]{97B7D7}} \color[HTML]{000000} 0.7348 & {\cellcolor[HTML]{B1C2DE}} \color[HTML]{000000} 0.6425 & {\cellcolor[HTML]{99B8D8}} \color[HTML]{000000} 0.7294 & {\cellcolor[HTML]{AFC1DD}} \color[HTML]{000000} 0.6557 & {\cellcolor[HTML]{94B6D7}} \color[HTML]{000000} 0.7453 & {\cellcolor[HTML]{CDD0E5}} \color[HTML]{000000} 0.5391 & {\cellcolor[HTML]{B8C6E0}} \color[HTML]{000000} 0.6181 & {\cellcolor[HTML]{A8BEDC}} \color[HTML]{000000} 0.6809 & {\cellcolor[HTML]{B0C2DE}} \color[HTML]{000000} 0.6473 & {\cellcolor[HTML]{A4BCDA}} \color[HTML]{000000} 0.6963 & {\cellcolor[HTML]{B9C6E0}} \color[HTML]{000000} 0.6137 \\
\bfseries hau & {\cellcolor[HTML]{A9BFDC}} \color[HTML]{000000} 0.6741 & {\cellcolor[HTML]{B0C2DE}} \color[HTML]{000000} 0.6493 & {\cellcolor[HTML]{B8C6E0}} \color[HTML]{000000} 0.6188 & {\cellcolor[HTML]{C6CCE3}} \color[HTML]{000000} 0.5637 & {\cellcolor[HTML]{D8D7E9}} \color[HTML]{000000} 0.4809 & {\cellcolor[HTML]{E7E3F0}} \color[HTML]{000000} 0.3897 & {\cellcolor[HTML]{BBC7E0}} \color[HTML]{000000} 0.6100 & {\cellcolor[HTML]{A9BFDC}} \color[HTML]{000000} 0.6724 & {\cellcolor[HTML]{93B5D6}} \color[HTML]{000000} 0.7488 & {\cellcolor[HTML]{C6CCE3}} \color[HTML]{000000} 0.5643 & {\cellcolor[HTML]{8FB4D6}} \color[HTML]{000000} 0.7623 & {\cellcolor[HTML]{DAD9EA}} \color[HTML]{000000} 0.4691 \\
\bfseries hin & {\cellcolor[HTML]{97B7D7}} \color[HTML]{000000} 0.7382 & {\cellcolor[HTML]{ADC1DD}} \color[HTML]{000000} 0.6586 & {\cellcolor[HTML]{ADC1DD}} \color[HTML]{000000} 0.6610 & {\cellcolor[HTML]{A7BDDB}} \color[HTML]{000000} 0.6830 & {\cellcolor[HTML]{A9BFDC}} \color[HTML]{000000} 0.6762 & {\cellcolor[HTML]{B9C6E0}} \color[HTML]{000000} 0.6118 & {\cellcolor[HTML]{A1BBDA}} \color[HTML]{000000} 0.7053 & {\cellcolor[HTML]{A7BDDB}} \color[HTML]{000000} 0.6842 & {\cellcolor[HTML]{ABBFDC}} \color[HTML]{000000} 0.6687 & {\cellcolor[HTML]{BDC8E1}} \color[HTML]{000000} 0.6009 & {\cellcolor[HTML]{A8BEDC}} \color[HTML]{000000} 0.6798 & {\cellcolor[HTML]{A8BEDC}} \color[HTML]{000000} 0.6795 \\
\bfseries ita & {\cellcolor[HTML]{C0C9E2}} \color[HTML]{000000} 0.5902 & {\cellcolor[HTML]{FFF7FB}} \color[HTML]{000000} - & {\cellcolor[HTML]{A7BDDB}} \color[HTML]{000000} 0.6859 & {\cellcolor[HTML]{96B6D7}} \color[HTML]{000000} 0.7392 & {\cellcolor[HTML]{88B1D4}} \color[HTML]{000000} 0.7882 & {\cellcolor[HTML]{FFF7FB}} \color[HTML]{000000} - & {\cellcolor[HTML]{FFF7FB}} \color[HTML]{000000} - & {\cellcolor[HTML]{FFF7FB}} \color[HTML]{000000} - & {\cellcolor[HTML]{FFF7FB}} \color[HTML]{000000} - & {\cellcolor[HTML]{FFF7FB}} \color[HTML]{000000} - & {\cellcolor[HTML]{FFF7FB}} \color[HTML]{000000} - & {\cellcolor[HTML]{FFF7FB}} \color[HTML]{000000} - \\
\bfseries khm & {\cellcolor[HTML]{ACC0DD}} \color[HTML]{000000} 0.6655 & {\cellcolor[HTML]{81AED2}} \color[HTML]{000000} 0.8094 & {\cellcolor[HTML]{A9BFDC}} \color[HTML]{000000} 0.6734 & {\cellcolor[HTML]{A4BCDA}} \color[HTML]{000000} 0.6940 & {\cellcolor[HTML]{B8C6E0}} \color[HTML]{000000} 0.6212 & {\cellcolor[HTML]{A5BDDB}} \color[HTML]{000000} 0.6924 & {\cellcolor[HTML]{A2BCDA}} \color[HTML]{000000} 0.6988 & {\cellcolor[HTML]{69A5CC}} \color[HTML]{000000} 0.8790 & {\cellcolor[HTML]{63A2CB}} \color[HTML]{000000} 0.8948 & {\cellcolor[HTML]{88B1D4}} \color[HTML]{000000} 0.7842 & {\cellcolor[HTML]{96B6D7}} \color[HTML]{000000} 0.7422 & {\cellcolor[HTML]{96B6D7}} \color[HTML]{000000} 0.7383 \\
\bfseries mya & {\cellcolor[HTML]{91B5D6}} \color[HTML]{000000} 0.7579 & {\cellcolor[HTML]{A5BDDB}} \color[HTML]{000000} 0.6894 & {\cellcolor[HTML]{ACC0DD}} \color[HTML]{000000} 0.6638 & {\cellcolor[HTML]{BDC8E1}} \color[HTML]{000000} 0.5986 & {\cellcolor[HTML]{A7BDDB}} \color[HTML]{000000} 0.6831 & {\cellcolor[HTML]{91B5D6}} \color[HTML]{000000} 0.7539 & {\cellcolor[HTML]{FFF7FB}} \color[HTML]{000000} - & {\cellcolor[HTML]{FFF7FB}} \color[HTML]{000000} - & {\cellcolor[HTML]{FFF7FB}} \color[HTML]{000000} - & {\cellcolor[HTML]{FFF7FB}} \color[HTML]{000000} - & {\cellcolor[HTML]{FFF7FB}} \color[HTML]{000000} - & {\cellcolor[HTML]{FFF7FB}} \color[HTML]{000000} - \\
\bfseries nep & {\cellcolor[HTML]{88B1D4}} \color[HTML]{000000} 0.7890 & {\cellcolor[HTML]{A4BCDA}} \color[HTML]{000000} 0.6932 & {\cellcolor[HTML]{9AB8D8}} \color[HTML]{000000} 0.7250 & {\cellcolor[HTML]{B4C4DF}} \color[HTML]{000000} 0.6352 & {\cellcolor[HTML]{67A4CC}} \color[HTML]{000000} 0.8812 & {\cellcolor[HTML]{86B0D3}} \color[HTML]{000000} 0.7895 & {\cellcolor[HTML]{88B1D4}} \color[HTML]{000000} 0.7844 & {\cellcolor[HTML]{80AED2}} \color[HTML]{000000} 0.8110 & {\cellcolor[HTML]{69A5CC}} \color[HTML]{000000} 0.8801 & {\cellcolor[HTML]{A7BDDB}} \color[HTML]{000000} 0.6829 & {\cellcolor[HTML]{A7BDDB}} \color[HTML]{000000} 0.6829 & {\cellcolor[HTML]{8BB2D4}} \color[HTML]{000000} 0.7784 \\
\bfseries ori & {\cellcolor[HTML]{ACC0DD}} \color[HTML]{000000} 0.6632 & {\cellcolor[HTML]{A2BCDA}} \color[HTML]{000000} 0.7023 & {\cellcolor[HTML]{93B5D6}} \color[HTML]{000000} 0.7530 & {\cellcolor[HTML]{8EB3D5}} \color[HTML]{000000} 0.7685 & {\cellcolor[HTML]{84B0D3}} \color[HTML]{000000} 0.7954 & {\cellcolor[HTML]{76AAD0}} \color[HTML]{000000} 0.8405 & {\cellcolor[HTML]{BBC7E0}} \color[HTML]{000000} 0.6083 & {\cellcolor[HTML]{ACC0DD}} \color[HTML]{000000} 0.6624 & {\cellcolor[HTML]{7BACD1}} \color[HTML]{000000} 0.8273 & {\cellcolor[HTML]{91B5D6}} \color[HTML]{000000} 0.7572 & {\cellcolor[HTML]{B8C6E0}} \color[HTML]{000000} 0.6170 & {\cellcolor[HTML]{A8BEDC}} \color[HTML]{000000} 0.6822 \\
\bfseries pan & {\cellcolor[HTML]{9FBAD9}} \color[HTML]{000000} 0.7115 & {\cellcolor[HTML]{A9BFDC}} \color[HTML]{000000} 0.6731 & {\cellcolor[HTML]{B9C6E0}} \color[HTML]{000000} 0.6152 & {\cellcolor[HTML]{A5BDDB}} \color[HTML]{000000} 0.6902 & {\cellcolor[HTML]{A1BBDA}} \color[HTML]{000000} 0.7031 & {\cellcolor[HTML]{C1CAE2}} \color[HTML]{000000} 0.5822 & {\cellcolor[HTML]{B1C2DE}} \color[HTML]{000000} 0.6467 & {\cellcolor[HTML]{91B5D6}} \color[HTML]{000000} 0.7561 & {\cellcolor[HTML]{B9C6E0}} \color[HTML]{000000} 0.6129 & {\cellcolor[HTML]{BDC8E1}} \color[HTML]{000000} 0.5963 & {\cellcolor[HTML]{ACC0DD}} \color[HTML]{000000} 0.6628 & {\cellcolor[HTML]{B1C2DE}} \color[HTML]{000000} 0.6422 \\
\bfseries pol & {\cellcolor[HTML]{9EBAD9}} \color[HTML]{000000} 0.7179 & {\cellcolor[HTML]{9AB8D8}} \color[HTML]{000000} 0.7253 & {\cellcolor[HTML]{B9C6E0}} \color[HTML]{000000} 0.6160 & {\cellcolor[HTML]{A1BBDA}} \color[HTML]{000000} 0.7058 & {\cellcolor[HTML]{B4C4DF}} \color[HTML]{000000} 0.6362 & {\cellcolor[HTML]{A4BCDA}} \color[HTML]{000000} 0.6929 & {\cellcolor[HTML]{FFF7FB}} \color[HTML]{000000} - & {\cellcolor[HTML]{FFF7FB}} \color[HTML]{000000} - & {\cellcolor[HTML]{FFF7FB}} \color[HTML]{000000} - & {\cellcolor[HTML]{FFF7FB}} \color[HTML]{000000} - & {\cellcolor[HTML]{FFF7FB}} \color[HTML]{000000} - & {\cellcolor[HTML]{FFF7FB}} \color[HTML]{000000} - \\
\bfseries rus & {\cellcolor[HTML]{A1BBDA}} \color[HTML]{000000} 0.7043 & {\cellcolor[HTML]{97B7D7}} \color[HTML]{000000} 0.7363 & {\cellcolor[HTML]{9EBAD9}} \color[HTML]{000000} 0.7156 & {\cellcolor[HTML]{B9C6E0}} \color[HTML]{000000} 0.6133 & {\cellcolor[HTML]{B0C2DE}} \color[HTML]{000000} 0.6507 & {\cellcolor[HTML]{ADC1DD}} \color[HTML]{000000} 0.6583 & {\cellcolor[HTML]{FFF7FB}} \color[HTML]{000000} - & {\cellcolor[HTML]{FFF7FB}} \color[HTML]{000000} - & {\cellcolor[HTML]{FFF7FB}} \color[HTML]{000000} - & {\cellcolor[HTML]{FFF7FB}} \color[HTML]{000000} - & {\cellcolor[HTML]{FFF7FB}} \color[HTML]{000000} - & {\cellcolor[HTML]{FFF7FB}} \color[HTML]{000000} - \\
\bfseries spa & {\cellcolor[HTML]{A2BCDA}} \color[HTML]{000000} 0.6996 & {\cellcolor[HTML]{8FB4D6}} \color[HTML]{000000} 0.7635 & {\cellcolor[HTML]{B8C6E0}} \color[HTML]{000000} 0.6197 & {\cellcolor[HTML]{C5CCE3}} \color[HTML]{000000} 0.5660 & {\cellcolor[HTML]{9CB9D9}} \color[HTML]{000000} 0.7219 & {\cellcolor[HTML]{A1BBDA}} \color[HTML]{000000} 0.7051 & {\cellcolor[HTML]{ACC0DD}} \color[HTML]{000000} 0.6639 & {\cellcolor[HTML]{A9BFDC}} \color[HTML]{000000} 0.6741 & {\cellcolor[HTML]{BFC9E1}} \color[HTML]{000000} 0.5960 & {\cellcolor[HTML]{A2BCDA}} \color[HTML]{000000} 0.6978 & {\cellcolor[HTML]{B0C2DE}} \color[HTML]{000000} 0.6511 & {\cellcolor[HTML]{8BB2D4}} \color[HTML]{000000} 0.7765 \\
\bfseries swa & {\cellcolor[HTML]{ACC0DD}} \color[HTML]{000000} 0.6668 & {\cellcolor[HTML]{C2CBE2}} \color[HTML]{000000} 0.5789 & {\cellcolor[HTML]{A5BDDB}} \color[HTML]{000000} 0.6913 & {\cellcolor[HTML]{B8C6E0}} \color[HTML]{000000} 0.6182 & {\cellcolor[HTML]{D1D2E6}} \color[HTML]{000000} 0.5206 & {\cellcolor[HTML]{CDD0E5}} \color[HTML]{000000} 0.5358 & {\cellcolor[HTML]{B0C2DE}} \color[HTML]{000000} 0.6472 & {\cellcolor[HTML]{B7C5DF}} \color[HTML]{000000} 0.6260 & {\cellcolor[HTML]{CACEE5}} \color[HTML]{000000} 0.5488 & {\cellcolor[HTML]{C2CBE2}} \color[HTML]{000000} 0.5789 & {\cellcolor[HTML]{BDC8E1}} \color[HTML]{000000} 0.5975 & {\cellcolor[HTML]{B8C6E0}} \color[HTML]{000000} 0.6174 \\
\bfseries tel & {\cellcolor[HTML]{91B5D6}} \color[HTML]{000000} 0.7573 & {\cellcolor[HTML]{B0C2DE}} \color[HTML]{000000} 0.6484 & {\cellcolor[HTML]{A7BDDB}} \color[HTML]{000000} 0.6851 & {\cellcolor[HTML]{B0C2DE}} \color[HTML]{000000} 0.6470 & {\cellcolor[HTML]{B0C2DE}} \color[HTML]{000000} 0.6514 & {\cellcolor[HTML]{A9BFDC}} \color[HTML]{000000} 0.6746 & {\cellcolor[HTML]{B1C2DE}} \color[HTML]{000000} 0.6419 & {\cellcolor[HTML]{93B5D6}} \color[HTML]{000000} 0.7517 & {\cellcolor[HTML]{99B8D8}} \color[HTML]{000000} 0.7323 & {\cellcolor[HTML]{B8C6E0}} \color[HTML]{000000} 0.6188 & {\cellcolor[HTML]{B1C2DE}} \color[HTML]{000000} 0.6460 & {\cellcolor[HTML]{B0C2DE}} \color[HTML]{000000} 0.6500 \\
\bfseries tur & {\cellcolor[HTML]{8EB3D5}} \color[HTML]{000000} 0.7660 & {\cellcolor[HTML]{94B6D7}} \color[HTML]{000000} 0.7450 & {\cellcolor[HTML]{A2BCDA}} \color[HTML]{000000} 0.7020 & {\cellcolor[HTML]{9FBAD9}} \color[HTML]{000000} 0.7119 & {\cellcolor[HTML]{B9C6E0}} \color[HTML]{000000} 0.6125 & {\cellcolor[HTML]{BCC7E1}} \color[HTML]{000000} 0.6019 & {\cellcolor[HTML]{8FB4D6}} \color[HTML]{000000} 0.7627 & {\cellcolor[HTML]{91B5D6}} \color[HTML]{000000} 0.7543 & {\cellcolor[HTML]{A1BBDA}} \color[HTML]{000000} 0.7033 & {\cellcolor[HTML]{94B6D7}} \color[HTML]{000000} 0.7438 & {\cellcolor[HTML]{C9CEE4}} \color[HTML]{000000} 0.5513 & {\cellcolor[HTML]{ADC1DD}} \color[HTML]{000000} 0.6579 \\
\bfseries urd & {\cellcolor[HTML]{A9BFDC}} \color[HTML]{000000} 0.6750 & {\cellcolor[HTML]{B0C2DE}} \color[HTML]{000000} 0.6480 & {\cellcolor[HTML]{BBC7E0}} \color[HTML]{000000} 0.6067 & {\cellcolor[HTML]{BDC8E1}} \color[HTML]{000000} 0.5986 & {\cellcolor[HTML]{B4C4DF}} \color[HTML]{000000} 0.6342 & {\cellcolor[HTML]{BFC9E1}} \color[HTML]{000000} 0.5915 & {\cellcolor[HTML]{B9C6E0}} \color[HTML]{000000} 0.6148 & {\cellcolor[HTML]{B9C6E0}} \color[HTML]{000000} 0.6162 & {\cellcolor[HTML]{BCC7E1}} \color[HTML]{000000} 0.6056 & {\cellcolor[HTML]{B4C4DF}} \color[HTML]{000000} 0.6341 & {\cellcolor[HTML]{B1C2DE}} \color[HTML]{000000} 0.6431 & {\cellcolor[HTML]{B3C3DE}} \color[HTML]{000000} 0.6381 \\
\bfseries zho & {\cellcolor[HTML]{73A9CF}} \color[HTML]{000000} 0.8543 & {\cellcolor[HTML]{91B5D6}} \color[HTML]{000000} 0.7551 & {\cellcolor[HTML]{89B1D4}} \color[HTML]{000000} 0.7795 & {\cellcolor[HTML]{A8BEDC}} \color[HTML]{000000} 0.6801 & {\cellcolor[HTML]{79ABD0}} \color[HTML]{000000} 0.8347 & {\cellcolor[HTML]{A5BDDB}} \color[HTML]{000000} 0.6882 & {\cellcolor[HTML]{8CB3D5}} \color[HTML]{000000} 0.7730 & {\cellcolor[HTML]{86B0D3}} \color[HTML]{000000} 0.7896 & {\cellcolor[HTML]{86B0D3}} \color[HTML]{000000} 0.7893 & {\cellcolor[HTML]{8CB3D5}} \color[HTML]{000000} 0.7724 & {\cellcolor[HTML]{7EADD1}} \color[HTML]{000000} 0.8162 & {\cellcolor[HTML]{96B6D7}} \color[HTML]{000000} 0.7412 \\
\hline
\end{tabular}
}
\caption{Per-language per-feature AUC performance of the appraisals-classifier system based on hold-out part of the train set.}
\label{tab:appraisals_auc}
\end{table*}

Regarding the comparison to other systems submitted to the shared task (see Table~\ref{tab:rank}), our system ranked in the top 20\% in 14 cases (out of 62 combinations of languages and subtasks). In 28 cases, the system ranked in the top 50\%. Only in a single case, it has performed the \textbf{1st} of all submitted systems (\textbf{Italian} in subtask 1). The best rank achieved in subtask 2 was \textbf{3rd in Nepali} and \textbf{4th in Urdu} in subtask 3. On the other hand, the system ranked in the bottom 20\% in 15 cases.

Based on the results, the manifestation identification seems to be the most challenging subtask for our submitted system. After the analysis of the individual monitored features (labels) of each subtask based on the released gold labels (after the system-submission deadline), we can see (Table~\ref{tab:performance_labels}) that ``other'' class in subtask 2 was the most challenging. In subtask 3, ``dehumanization'', ``lack of empathy'', and ``invalidation'' are particularly misclassified.

\subsection{Appraisals for Polarization Detection}
\label{ssec:app-polar}

As the results in Table~\ref{tab:appraisals_performance} indicate, the appraisals can be used also for polarization detection. Such cognitive signals capture features that are important for polarization as well. Since the performance for subtasks 2 and 3 is rather random, we have looked at the AUC values per each label, showing a threshold-independent classification capability. The results are shown in Table~\ref{tab:appraisals_auc}. Such results clearly indicate a high potential of using (multilingual) appraisals for polarization-related indicators, which might be worthy exploring further in future work.

\begin{table}[!t]
\centering
\resizebox{0.95\linewidth}{!}{
\begin{tabular}{r|c|c|c}
\hline
 & \multicolumn{3}{c}{\textbf{Macro F1}}\\
\bfseries Language & \bfseries Subtask 1 & \bfseries Subtask 2 & \bfseries Subtask 3 \\
\hline
\bfseries amh & {\cellcolor[HTML]{C8CDE4}} \color[HTML]{000000} 0.5566 & {\cellcolor[HTML]{D6D6E9}} \color[HTML]{000000} 0.4897 & {\cellcolor[HTML]{D3D4E7}} \color[HTML]{000000} 0.5062 \\
\bfseries arb & {\cellcolor[HTML]{B4C4DF}} \color[HTML]{000000} 0.6338 & {\cellcolor[HTML]{DBDAEB}} \color[HTML]{000000} 0.4633 & {\cellcolor[HTML]{D6D6E9}} \color[HTML]{000000} 0.4924 \\
\bfseries ben & {\cellcolor[HTML]{A7BDDB}} \color[HTML]{000000} 0.6841 & {\cellcolor[HTML]{D2D2E7}} \color[HTML]{000000} 0.5168 & {\cellcolor[HTML]{D6D6E9}} \color[HTML]{000000} 0.4919 \\
\bfseries deu & {\cellcolor[HTML]{B8C6E0}} \color[HTML]{000000} 0.6188 & {\cellcolor[HTML]{D7D6E9}} \color[HTML]{000000} 0.4869 & {\cellcolor[HTML]{DDDBEC}} \color[HTML]{000000} 0.4511 \\
\bfseries eng & {\cellcolor[HTML]{B1C2DE}} \color[HTML]{000000} 0.6464 & {\cellcolor[HTML]{D2D2E7}} \color[HTML]{000000} 0.5174 & {\cellcolor[HTML]{D9D8EA}} \color[HTML]{000000} 0.4744 \\
\bfseries fas & {\cellcolor[HTML]{D4D4E8}} \color[HTML]{000000} 0.5040 & {\cellcolor[HTML]{D1D2E6}} \color[HTML]{000000} 0.5217 & {\cellcolor[HTML]{D8D7E9}} \color[HTML]{000000} 0.4801 \\
\bfseries hau & {\cellcolor[HTML]{D9D8EA}} \color[HTML]{000000} 0.4718 & {\cellcolor[HTML]{D6D6E9}} \color[HTML]{000000} 0.4939 & {\cellcolor[HTML]{D6D6E9}} \color[HTML]{000000} 0.4945 \\
\bfseries hin & {\cellcolor[HTML]{D7D6E9}} \color[HTML]{000000} 0.4844 & {\cellcolor[HTML]{D2D3E7}} \color[HTML]{000000} 0.5113 & {\cellcolor[HTML]{C2CBE2}} \color[HTML]{000000} 0.5775 \\
\bfseries ita & {\cellcolor[HTML]{E0DEED}} \color[HTML]{000000} 0.4335 & {\cellcolor[HTML]{DBDAEB}} \color[HTML]{000000} 0.4615 & {\cellcolor[HTML]{FFF7FB}} \color[HTML]{000000} - \\
\bfseries khm & {\cellcolor[HTML]{D9D8EA}} \color[HTML]{000000} 0.4759 & {\cellcolor[HTML]{D2D2E7}} \color[HTML]{000000} 0.5196 & {\cellcolor[HTML]{D4D4E8}} \color[HTML]{000000} 0.5018 \\
\bfseries mya & {\cellcolor[HTML]{A4BCDA}} \color[HTML]{000000} 0.6944 & {\cellcolor[HTML]{D2D2E7}} \color[HTML]{000000} 0.5189 & {\cellcolor[HTML]{FFF7FB}} \color[HTML]{000000} - \\
\bfseries nep & {\cellcolor[HTML]{9FBAD9}} \color[HTML]{000000} 0.7107 & {\cellcolor[HTML]{D9D8EA}} \color[HTML]{000000} 0.4770 & {\cellcolor[HTML]{C9CEE4}} \color[HTML]{000000} 0.5525 \\
\bfseries ori & {\cellcolor[HTML]{DCDAEB}} \color[HTML]{000000} 0.4563 & {\cellcolor[HTML]{D9D8EA}} \color[HTML]{000000} 0.4788 & {\cellcolor[HTML]{D8D7E9}} \color[HTML]{000000} 0.4820 \\
\bfseries pan & {\cellcolor[HTML]{ABBFDC}} \color[HTML]{000000} 0.6673 & {\cellcolor[HTML]{D6D6E9}} \color[HTML]{000000} 0.4903 & {\cellcolor[HTML]{D7D6E9}} \color[HTML]{000000} 0.4854 \\
\bfseries pol & {\cellcolor[HTML]{ADC1DD}} \color[HTML]{000000} 0.6581 & {\cellcolor[HTML]{D3D4E7}} \color[HTML]{000000} 0.5086 & {\cellcolor[HTML]{FFF7FB}} \color[HTML]{000000} - \\
\bfseries rus & {\cellcolor[HTML]{C8CDE4}} \color[HTML]{000000} 0.5566 & {\cellcolor[HTML]{D8D7E9}} \color[HTML]{000000} 0.4811 & {\cellcolor[HTML]{FFF7FB}} \color[HTML]{000000} - \\
\bfseries spa & {\cellcolor[HTML]{B1C2DE}} \color[HTML]{000000} 0.6440 & {\cellcolor[HTML]{D6D6E9}} \color[HTML]{000000} 0.4907 & {\cellcolor[HTML]{D9D8EA}} \color[HTML]{000000} 0.4786 \\
\bfseries swa & {\cellcolor[HTML]{B5C4DF}} \color[HTML]{000000} 0.6268 & {\cellcolor[HTML]{D3D4E7}} \color[HTML]{000000} 0.5074 & {\cellcolor[HTML]{D9D8EA}} \color[HTML]{000000} 0.4760 \\
\bfseries tel & {\cellcolor[HTML]{A2BCDA}} \color[HTML]{000000} 0.7005 & {\cellcolor[HTML]{DDDBEC}} \color[HTML]{000000} 0.4534 & {\cellcolor[HTML]{DBDAEB}} \color[HTML]{000000} 0.4640 \\
\bfseries tur & {\cellcolor[HTML]{A2BCDA}} \color[HTML]{000000} 0.6997 & {\cellcolor[HTML]{D1D2E6}} \color[HTML]{000000} 0.5217 & {\cellcolor[HTML]{C5CCE3}} \color[HTML]{000000} 0.5667 \\
\bfseries urd & {\cellcolor[HTML]{DBDAEB}} \color[HTML]{000000} 0.4625 & {\cellcolor[HTML]{C6CCE3}} \color[HTML]{000000} 0.5611 & {\cellcolor[HTML]{CDD0E5}} \color[HTML]{000000} 0.5396 \\
\bfseries zho & {\cellcolor[HTML]{89B1D4}} \color[HTML]{000000} 0.7803 & {\cellcolor[HTML]{CDD0E5}} \color[HTML]{000000} 0.5382 & {\cellcolor[HTML]{D1D2E6}} \color[HTML]{000000} 0.5243 \\
\hline
\end{tabular}
}
\caption{Per-language performance of the appraisals-classifier system based on hold-out part of the train set.}
\label{tab:appraisals_performance}
\end{table}

\section{Conclusion}

The submitted multilingual system has shown that a single system can handle all languages combined, competitively in comparison to other systems. However, the performance differs among languages, and for some the approach is not suitable. The multi-label classification subtasks have been more challenging for the system than the binary classification, for which the original system has been tuned. In future, the system might be modified to be finetuned individually for each language or some group or similar languages (not all of them). Furthermore, the appraisal-based alternative and its combination with the pure finetuned detector seem worthy to explore.

\section*{Limitations}

We have explored only small set of base language models. Others could be better performing, at least for some languages. We have tested only the languages included in the shared task, thus the generalization to other languages is unevaluated. We have limited the training set of samples only to the official train and dev splits, available in the shared task. Other publicly available datasets could be used for training as well.

\section*{Acknowledgments}
This work was partially supported by \textit{LorAI -- Low Resource Artificial Intelligence}, a project funded by Horizon Europe under \href{https://doi.org/10.3030/101136646}{GA No.101136646} and partially by the EU NextGenerationEU through the Recovery and Resilience Plan for Slovakia under the project No. 09I01-03-V04-00068.

\textbf{Computational resources}. This work was supported by the use of computational resources of the supercomputer PERUN, operated by the Supercomputing Centre at the Technical University of Košice (TUKE), Slovakia with the support of the European Union from the funds of the Recovery and Resilience Plan of the Slovak Republic within the framework of project No. 17I03-04-P03-00001, Development and design of a supercomputer for the National Supercomputing Center. We also acknowledge EuroHPC Joint Undertaking for awarding us access to Leonardo at CINECA, Italy.

\bibliography{custom, anthology}

@inproceedings{naseem-etal-2026-polar,

  title     = {{S}em{E}val-2026 Task 9: Detecting Multilingual, Multicultural and Multievent Online Polarization},

  author    = {Naseem, Usman and Geislinger, Robert and Ren, Juan and Kohail, Sarah and Garrido Veliz, Rudy and Sam Sahil, P and Zhang, Yiran and Stranisci, Marco Antonio and Abdulmumin, Idris and Alacam, {"O}zge and Acar{"u}rk, Cengiz and Jabr, Aisha and Anwar, Saba and Ayele, Abinew Ali and Tutubalina, Elena and Htet, Aung Kyaw and Wang, Xintong and Thapa, Surendrabikram and Chakraborty, Tanmoy and Kodati, Dheeraj and Moradizeyveh, Sahar and Alam, Firoj and Thu, Ye Kyaw and Parida, Shantipriya and Qazi, Ihsan Ayyub and Onyango, Nelson Odhiambo and Siro, Clemencia and Ahmad, Ibrahim Said and Wanzare, Lilian and Ali, Adem Chanie and Semmann, Martin and Biemann, Chris and Muhammad, Shamsuddeen Hassan and Yimam, Seid Muhie},

  booktitle = {Proceedings of the 20th International Workshop on Semantic Evaluation (SemEval-2026)},

  year      = {2026},

  publisher = {Association for Computational Linguistics},

}

@misc{naseem2026polarbenchmarkmultilingualmulticultural,

      title={POLAR: A Benchmark for Multilingual, Multicultural, and Multi-Event Online Polarization},

      author={Usman Naseem and Robert Geislinger and Juan Ren and Sarah Kohail and Rudy Garrido Veliz and P Sam Sahil and Yiran Zhang and Marco Antonio Stranisci and Idris Abdulmumin and Özge Alacam and Cengiz Acartürk and Aisha Jabr and Saba Anwar and Abinew Ali Ayele and Simona Frenda and Alessandra Teresa Cignarella and Elena Tutubalina and Oleg Rogov and Aung Kyaw Htet and Xintong Wang and Surendrabikram Thapa and Kritesh Rauniyar and Tanmoy Chakraborty and Arfeen Zeeshan and Dheeraj Kodati and Satya Keerthi and Sahar Moradizeyveh and Firoj Alam and Arid Hasan and Syed Ishtiaque Ahmed and Ye Kyaw Thu and Shantipriya Parida and Ihsan Ayyub Qazi and Lilian Wanzare and Nelson Odhiambo Onyango and Clemencia Siro and Jane Wanjiru Kimani and Ibrahim Said Ahmad and Adem Chanie Ali and Martin Semmann and Chris Biemann and Shamsuddeen Hassan Muhammad and Seid Muhie Yimam},

      year={2026},

      eprint={2505.20624},

      archivePrefix={arXiv},

      primaryClass={cs.CL},

      url={https://arxiv.org/abs/2505.20624},

}

@misc{macko2025increasingrobustnessfinetunedmultilingual,
      title={Increasing the Robustness of the Fine-tuned Multilingual Machine-Generated Text Detectors}, 
      author={Dominik Macko and Robert Moro and Ivan Srba},
      year={2025},
      eprint={2503.15128},
      archivePrefix={arXiv},
      primaryClass={cs.CL},
      url={https://arxiv.org/abs/2503.15128}, 
}

@inproceedings{Bevendorff2025OverviewOT,
  title={Overview of the "Voight-Kampff" Generative {AI} Authorship Verification Task at {PAN} and {ELOQUENT} 2025},
  author={Janek Bevendorff and Yuxia Wang and Jussi Karlgren and Matti Wiegmann and Maik Fr{\"o}be and Akim Tsivgun and Jinyan Su and Zhuohan Xie and Mervat T. Abassy and Jonibek Mansurov and Rui Xing and Minh Ngoc Ta and Kareem Ashraf Elozeiri and Tianle Gu and Raj Vardhan Tomar and Jiahui Geng and Ekaterina Artemova and Artem Shelmanov and Nizar Habash and Efstathios Stamatatos and Iryna Gurevych and Preslav Nakov and Martin Potthast and Benno Stein},
  booktitle = {Working Notes of {CLEF} 2025 -- Conference and Labs of the Evaluation Forum, {CEUR-WS.org}},
  year = 2025,
  url = {https://ceur-ws.org/Vol-4038/paper_277.pdf}
}

@inproceedings{mdok,
    author = {Dominik Macko},
    title = {mdok of {KInIT}: Robustly Fine-tuned {LLM} for Binary and Multiclass {AI}-Generated Text Detection},
    booktitle = {Working Notes of {CLEF} 2025 -- Conference and Labs of the Evaluation Forum, {CEUR-WS.org}},
    year = 2025,
    url = {https://ceur-ws.org/Vol-4038/paper_307.pdf}
}

@misc{lacava2025authorshipattributionmultilingualmachinegenerated,
      title={Authorship Attribution in Multilingual Machine-Generated Texts}, 
      author={Lucio La Cava and Dominik Macko and Róbert Móro and Ivan Srba and Andrea Tagarelli},
      year={2025},
      eprint={2508.01656},
      archivePrefix={arXiv},
      primaryClass={cs.CL},
      url={https://arxiv.org/abs/2508.01656}, 
}

@inproceedings{NEURIPS2023_1feb8787,
 author = {Dettmers, Tim and Pagnoni, Artidoro and Holtzman, Ari and Zettlemoyer, Luke},
 booktitle = {Advances in Neural Information Processing Systems},
 editor = {A. Oh and T. Naumann and A. Globerson and K. Saenko and M. Hardt and S. Levine},
 pages = {10088--10115},
 publisher = {Curran Associates, Inc.},
 title = {{QLoRA}: Efficient Finetuning of Quantized {LLMs}},
 url = {https://proceedings.neurips.cc/paper_files/paper/2023/file/1feb87871436031bdc0f2beaa62a049b-Paper-Conference.pdf},
 volume = {36},
 year = {2023}
}

@misc{gemmateam2025gemma3technicalreport,
      title={Gemma 3 Technical Report}, 
      author={Gemma Team and Aishwarya Kamath and Johan Ferret and Shreya Pathak and Nino Vieillard and Ramona Merhej and Sarah Perrin and Tatiana Matejovicova and Alexandre Ramé and Morgane Rivière and Louis Rouillard and Thomas Mesnard and Geoffrey Cideron and Jean-bastien Grill and Sabela Ramos and Edouard Yvinec and Michelle Casbon and Etienne Pot and Ivo Penchev and Gaël Liu and Francesco Visin and Kathleen Kenealy and Lucas Beyer and Xiaohai Zhai and Anton Tsitsulin and Robert Busa-Fekete and Alex Feng and Noveen Sachdeva and Benjamin Coleman and Yi Gao and Basil Mustafa and Iain Barr and Emilio Parisotto and David Tian and Matan Eyal and Colin Cherry and Jan-Thorsten Peter and Danila Sinopalnikov and Surya Bhupatiraju and Rishabh Agarwal and Mehran Kazemi and Dan Malkin and Ravin Kumar and David Vilar and Idan Brusilovsky and Jiaming Luo and Andreas Steiner and Abe Friesen and Abhanshu Sharma and Abheesht Sharma and Adi Mayrav Gilady and Adrian Goedeckemeyer and Alaa Saade and Alex Feng and Alexander Kolesnikov and Alexei Bendebury and Alvin Abdagic and Amit Vadi and András György and André Susano Pinto and Anil Das and Ankur Bapna and Antoine Miech and Antoine Yang and Antonia Paterson and Ashish Shenoy and Ayan Chakrabarti and Bilal Piot and Bo Wu and Bobak Shahriari and Bryce Petrini and Charlie Chen and Charline Le Lan and Christopher A. Choquette-Choo and CJ Carey and Cormac Brick and Daniel Deutsch and Danielle Eisenbud and Dee Cattle and Derek Cheng and Dimitris Paparas and Divyashree Shivakumar Sreepathihalli and Doug Reid and Dustin Tran and Dustin Zelle and Eric Noland and Erwin Huizenga and Eugene Kharitonov and Frederick Liu and Gagik Amirkhanyan and Glenn Cameron and Hadi Hashemi and Hanna Klimczak-Plucińska and Harman Singh and Harsh Mehta and Harshal Tushar Lehri and Hussein Hazimeh and Ian Ballantyne and Idan Szpektor and Ivan Nardini and Jean Pouget-Abadie and Jetha Chan and Joe Stanton and John Wieting and Jonathan Lai and Jordi Orbay and Joseph Fernandez and Josh Newlan and Ju-yeong Ji and Jyotinder Singh and Kat Black and Kathy Yu and Kevin Hui and Kiran Vodrahalli and Klaus Greff and Linhai Qiu and Marcella Valentine and Marina Coelho and Marvin Ritter and Matt Hoffman and Matthew Watson and Mayank Chaturvedi and Michael Moynihan and Min Ma and Nabila Babar and Natasha Noy and Nathan Byrd and Nick Roy and Nikola Momchev and Nilay Chauhan and Noveen Sachdeva and Oskar Bunyan and Pankil Botarda and Paul Caron and Paul Kishan Rubenstein and Phil Culliton and Philipp Schmid and Pier Giuseppe Sessa and Pingmei Xu and Piotr Stanczyk and Pouya Tafti and Rakesh Shivanna and Renjie Wu and Renke Pan and Reza Rokni and Rob Willoughby and Rohith Vallu and Ryan Mullins and Sammy Jerome and Sara Smoot and Sertan Girgin and Shariq Iqbal and Shashir Reddy and Shruti Sheth and Siim Põder and Sijal Bhatnagar and Sindhu Raghuram Panyam and Sivan Eiger and Susan Zhang and Tianqi Liu and Trevor Yacovone and Tyler Liechty and Uday Kalra and Utku Evci and Vedant Misra and Vincent Roseberry and Vlad Feinberg and Vlad Kolesnikov and Woohyun Han and Woosuk Kwon and Xi Chen and Yinlam Chow and Yuvein Zhu and Zichuan Wei and Zoltan Egyed and Victor Cotruta and Minh Giang and Phoebe Kirk and Anand Rao and Kat Black and Nabila Babar and Jessica Lo and Erica Moreira and Luiz Gustavo Martins and Omar Sanseviero and Lucas Gonzalez and Zach Gleicher and Tris Warkentin and Vahab Mirrokni and Evan Senter and Eli Collins and Joelle Barral and Zoubin Ghahramani and Raia Hadsell and Yossi Matias and D. Sculley and Slav Petrov and Noah Fiedel and Noam Shazeer and Oriol Vinyals and Jeff Dean and Demis Hassabis and Koray Kavukcuoglu and Clement Farabet and Elena Buchatskaya and Jean-Baptiste Alayrac and Rohan Anil and Dmitry and Lepikhin and Sebastian Borgeaud and Olivier Bachem and Armand Joulin and Alek Andreev and Cassidy Hardin and Robert Dadashi and Léonard Hussenot},
      year={2025},
      eprint={2503.19786},
      archivePrefix={arXiv},
      primaryClass={cs.CL},
      url={https://arxiv.org/abs/2503.19786}, 
}

@misc{yang2025qwen3technicalreport,
      title={Qwen3 Technical Report}, 
      author={An Yang and Anfeng Li and Baosong Yang and Beichen Zhang and Binyuan Hui and Bo Zheng and Bowen Yu and Chang Gao and Chengen Huang and Chenxu Lv and Chujie Zheng and Dayiheng Liu and Fan Zhou and Fei Huang and Feng Hu and Hao Ge and Haoran Wei and Huan Lin and Jialong Tang and Jian Yang and Jianhong Tu and Jianwei Zhang and Jianxin Yang and Jiaxi Yang and Jing Zhou and Jingren Zhou and Junyang Lin and Kai Dang and Keqin Bao and Kexin Yang and Le Yu and Lianghao Deng and Mei Li and Mingfeng Xue and Mingze Li and Pei Zhang and Peng Wang and Qin Zhu and Rui Men and Ruize Gao and Shixuan Liu and Shuang Luo and Tianhao Li and Tianyi Tang and Wenbiao Yin and Xingzhang Ren and Xinyu Wang and Xinyu Zhang and Xuancheng Ren and Yang Fan and Yang Su and Yichang Zhang and Yinger Zhang and Yu Wan and Yuqiong Liu and Zekun Wang and Zeyu Cui and Zhenru Zhang and Zhipeng Zhou and Zihan Qiu},
      year={2025},
      eprint={2505.09388},
      archivePrefix={arXiv},
      primaryClass={cs.CL},
      url={https://arxiv.org/abs/2505.09388}, 
}

@article{scherer1994evidence,
title={Evidence for universality and cultural variation of differential emotion response patterning},
author={Scherer, Klaus R and Wallbott, Harald G},
journal={Journal of personality and social psychology},
volume={66},
number={2},
pages={310},
year={1994},
publisher={American Psychological Association}
}

@inproceedings{conneau2020unsupervised,
  title={Unsupervised Cross-lingual Representation Learning at Scale},
  author={Conneau, Alexis and Khandelwal, Kartikay and Goyal, Naman and Chaudhary, Vishrav and Wenzek, Guillaume and Guzm{\'a}n, Francisco and Grave, Edouard and Ott, Myle and Zettlemoyer, Luke and Stoyanov, Veselin},
  booktitle={Proceedings of the 58th Annual Meeting of the Association for Computational Linguistics},
  pages={8440--8451},
  year={2020}
}

@inproceedings{troiano2019crowdsourcing,
  title={Crowdsourcing and validating event-focused emotion corpora for German and English},
  author={Troiano, Enrica and Pad{\'o}, Sebastian and Klinger, Roman},
  booktitle={Proceedings of the 57th Annual Meeting of the Association for Computational Linguistics},
  pages={4005--4011},
  year={2019}
}

\appendix

\section{Computational Resources}
\label{sec:A}

For experiments regarding model finetuning and inference processes, we have used multiple GPU-accelerated systems. Experiments using $1\times$ NVIDIA A100 64GB GPU taken around 200 GPU hours, and using $1\times$ NVIDIA H200 140GB GPU taken around 100 GPU hours. Analysis has been done without the GPU acceleration.

\section{Appraisal Annotations and Polarization}
\label{sec:B}

This appendix summarizes the datasets and architecture of the appraisal estimation framework underlying the empirical results reported in the main paper.

\paragraph{Theoretical basis.}
Appraisal theory is a cognitive and evolutionary account of emotion, positing that emotional experiences arise from evaluations of surrounding stimuli against context-relevant criteria such as \textit{pleasantness}, \textit{urgency}, and \textit{alignment with social norms}. These appraisal dimensions offer a richer substrate for understanding affective-cognitive links than coarse categorical emotion labels alone.

\paragraph{Corpus.}
The \textit{International Survey on Emotion Antecedents and Reactions} (ISEAR;~\citealt{scherer1994evidence}) corpus comprises English-language first-person descriptions of emotional experiences across seven categories: anger, disgust, fear, guilt, joy, shame, and sadness. Its emphasis on subjective experiential accounts distinguishes it from other NLP emotion resources. Building on this, \citet{troiano2019crowdsourcing} produced \textit{deISEAR} and \textit{enISEAR} --- crowdsourced, emotion-annotated datasets that preserve ISEAR's design principles while surfacing divergences between experiencer and reader emotion judgements.

\paragraph{Annotation scheme.}
From these corpora, appraisal objectives, including \textit{self-consequences}, \textit{consequences for others}, and \textit{situational control}, were annotated to capture whether the experiencer attributed their emotional state to each factor. These appraisal-level annotations, rather than emotion category labels, provide grounded cues to the link between event cognition and affective language. This monolingual, dimensionally richer appraisal mechanism has also proven useful in adjacent tasks such as emotion recognition in conversation~\citep{debnath-etal-2025-appraisal}.

\paragraph{Architecture.}
Appraisal estimation is implemented as a multitask binary classifier trained with weighted binary cross-entropy loss, designed to minimize appraisal mischaracterization. The resulting appraisal estimates are used as supplemental embeddings for the polarization detection and SemEval subtask objectives, augmenting the primary representation with affective-cognitive signals.

\paragraph{Findings and outlook.}
Incorporating deISEAR and enISEAR appraisals as additional embeddings yields promising but mixed results for polarization detection. Key limitations include the specificity of the selected appraisal dimensions, the method of appraisal representation learning and integration, and limited multilingual coverage. Nevertheless, performance improvements on subtasks with stronger affective-cognitive content (such as dehumanization, lack of empathy, and related emotional expression markers) suggest meaningful potential for future work exploiting these signals in text and conversation.

\end{document}